\pdfoutput=1

\documentclass[11pt]{article}

\usepackage[preprint]{acl}

\usepackage{times}
\usepackage{latexsym}

\usepackage[T1]{fontenc}

\usepackage[utf8]{inputenc}

\usepackage{microtype}

\usepackage{inconsolata}

\usepackage{graphicx}
\usepackage{placeins}

\usepackage[symbol]{footmisc}

\newcommand{\methodl}{Branch-and-Merge\xspace}
\newcommand{\method}{\textsc{BaM}\xspace}
\newcommand{\task}{language\xspace}

\newcommand{\cold}{\textsc{ColD Fusion}\xspace}

\newcommand{\lora}{\textsc{LoRA}\xspace}
\newcommand{\slerp}{\textsc{Slerp}\xspace}
\newcommand{\merge}{\textsc{Merge}\xspace}
\newcommand{\ties}{\textsc{Ties}\xspace}
\newcommand{\linear}{\textsc{Linear}\xspace}
\newcommand{\dares}{\textsc{Dares}\xspace}
\newcommand{\modelar}{\textsc{Task Arithmetic}\xspace}
\newcommand{\modelbread}{\textsc{Model Breadcrumbs}\xspace}
\newcommand{\modelstock}{\textsc{Model Stock}\xspace}

\usepackage{mathtools}
\usepackage{amsfonts}
\usepackage{dsfont}
\usepackage{xspace}
\usepackage{siunitx}
\usepackage{booktabs}       
\usepackage[capitalise]{cleveref}
\usepackage{multirow}

\usepackage{algorithm}
\usepackage[noend]{algorithmic}

\newcolumntype{x}[2]{S[table-format=#1.#2,table-auto-round]}

\usepackage{pifont}
\newcommand{\cmark}{\ding{51}}%
\newcommand{\xmark}{\ding{55}}%

\newcommand{\bc}[1]{\mathcal{#1}}

\newcommand{\mbf}[1]{\mathbf{#1}}

\NewDocumentCommand{\phimodel}{O{}}{%
\textsc{Phi}\ifstrempty{#1}{}{-{#1}}\xspace
}

\NewDocumentCommand{\gemma}{O{}}{%
\textsc{Gemma}\ifstrempty{#1}{}{-{#1}}\xspace
}

\NewDocumentCommand{\llama}{O{}O{}}{%
\textsc{Llama}\ifstrempty{#1}{}{-{#1}}\ifstrempty{#2}{}{-{#2}}\xspace
}

\NewDocumentCommand{\llamainstruct}{O{}O{}}{%
\textsc{Llama}\ifstrempty{#1}{}{-{#1}}-\textsc{Instruct}\ifstrempty{#2}{}{-{#2}}\xspace
}

\NewDocumentCommand{\gpt}{O{}}{%
\textsc{GPT}\ifstrempty{#1}{}{-{#1}}\xspace
}

\NewDocumentCommand{\gptturbo}{O{}}{%
\textsc{GPT}\ifstrempty{#1}{}{-{#1}}-\textsc{Turbo}\xspace
}

\NewDocumentCommand{\mistral}{O{}O{}}{%
\textsc{Mistral}\ifstrempty{#1}{}{-{#1}}\ifstrempty{#2}{}{-v0.{#2}}\xspace
}

\NewDocumentCommand{\mistralinstruct}{O{}O{}}{%
\textsc{Mistral-Instruct}\ifstrempty{#1}{}{-{#1}}\ifstrempty{#2}{}{-v0.{#2}}\xspace
}

\usepackage{tikz}

\usetikzlibrary{arrows}
\usetikzlibrary{automata}
\usetikzlibrary{calc}
\usetikzlibrary{backgrounds}
\usetikzlibrary{decorations.markings}
\usetikzlibrary{decorations.pathmorphing}
\usetikzlibrary{decorations.pathreplacing}
\usetikzlibrary{fit}
\usetikzlibrary{patterns}
\usetikzlibrary{positioning}
\usetikzlibrary{shadows}
\usetikzlibrary{shapes}
\usetikzlibrary{shapes.geometric}
\usetikzlibrary{arrows.meta, calc}

\definecolor{mypurple}{HTML}{613F99}
\definecolor{myblue}{HTML}{0071BC}
\definecolor{mygreen}{HTML}{3C8031}

\newcommand{\boxb}{\protect\tikz[]{\node[fill, aspect=1, color=myblue!40, inner sep=0pt, minimum size=2.1mm]{};}\xspace}
\newcommand{\boxg}{\protect\tikz[]{\node[fill, aspect=1, color=mygreen!40, inner sep=0pt, minimum size=2.1mm]{};}\xspace}
\newcommand{\boxp}{\protect\tikz[]{\node[fill, aspect=1, color=mypurple!40, inner sep=0pt, minimum size=2.1mm]{};}\xspace}

\newcommand{\markerx}[3]{\draw[-, line width=2.8pt, color=#3]($(#1) + (-#2,-#2)$) -- ($(#1) + (#2,#2)$) ($(#1) + (-#2,+#2)$) -- ($(#1) + (+#2,-#2)$);}

\title{Mitigating Catastrophic Forgetting in Language Transfer \\via Model Merging}

\author{\hspace{-1em}Anton Alexandrov\footnotemark$^{1}$, Veselin Raychev$^{1,2}$, Mark Niklas Müller$^{2,3}$\\ 
        \textbf{Ce Zhang$^{1,4,5}$, Martin Vechev$^{1,3}$, Kristina Toutanova$^{1,6}$}\\
	\hspace{-3em}$^{1}$ INSAIT, Sofia University ``St. Kliment Ohridski''\hspace{3em} $^{2}$LogicStar.ai\\ \hspace{-3em}$^{3}$ ETH Zurich\hspace{1em}$^{4}$ University of Chicago\hspace{1em}$^{5}$ Together AI\hspace{1em} $^{6}$ Google DeepMind\hspace{0.5em} \hfil\\
}

\begin{document}
\maketitle
\begin{abstract}
As open-weight large language models (LLMs) achieve ever more impressive performances across a wide range of tasks in English, practitioners aim to adapt these models to different languages. However, such language adaptation is often accompanied by catastrophic forgetting of the base model's capabilities, severely limiting the usefulness of the resulting model.
We address this issue by proposing \methodl (\method), a new adaptation method based on iteratively merging multiple models, fine-tuned on a subset of the available training data. \method is based on the insight that this yields lower magnitude but higher quality weight changes, reducing forgetting of the source domain while maintaining learning on the target domain. We demonstrate in an extensive empirical study on Bulgarian and German that \method can significantly reduce forgetting while matching or even improving target domain performance compared to both standard continued pretraining and instruction finetuning across different model architectures.
    \end{abstract}

\section{Introduction}
Large language models have shown remarkable capabilities, particularly in English. However, for less prevalent languages, performance can be significantly lower, making additional adaptation paramount \citep{ZhaoZGZGH24,CuiY24}. 
\footnotetext[0]{$^*$ Correspondence author: anton.alexandrov@insait.ai}

\paragraph{Catastrophic Forgetting} Unfortunately, most adaptation techniques come at the cost of catastrophic forgetting of the base model's capabilities \citep{ZhaiTLCQLM23,ShiXWQWWW24,LiL24,GogoulouLBN23}.
At the same time, retaining these capabilities is often crucial for solving downstream tasks in a new \task. For example, math and coding skills learned in English can be extremely helpful for general problem-solving or reasoning tasks in other languages.

\begin{figure}
    \centering
    \resizebox{\linewidth}{!}{
    \begin{tikzpicture}[node distance=1cm and 1cm, every node/.style={font=\sffamily}]

    \def\dy{1.3cm}
    \def\ddx{0.125cm}
    \def\ddy{0.25cm}

    \def\dwidth{1.1cm}
    \def\dheight{0.8cm}
    \def\boxwidth{1.6cm}

    \node (main) [rounded corners, thick, minimum width=5.5cm, minimum height=4.8cm, fill=gray!10] at (0,0) {};

    \node at ($(main.north) + (0,-0.3cm)$) {\large\bf \methodl};

    \node (base) [rounded corners, minimum height=\dy, minimum width=\boxwidth, fill=gray!30, anchor=east, align=center] at ($(main.west) + (-0.6, 0.7)$) {\bf Base \\\bf Model};

    \node (TrainA) [rounded corners, minimum height=\dheight, minimum width=2.6cm, fill=mygreen!40, anchor=west, align=center] at ($(main.west |- base.east) + (0.5, 0.7)$) {\bf Train on $\mathcal{X}_{i+1}$};

    \node (TrainB) [rounded corners, minimum height=\dheight, minimum width=2.6cm, fill=mygreen!20, anchor=west, align=center] at ($(TrainA.west |- base.east) + (0.0, -0.7)$) {\bf Train on $\mathcal{X}_{i}$};

    \draw[-{Triangle},draw=black, line width=1pt] (base.east) -- ($(main.west |- base.east) + (0.2cm,0.0cm)$) -- ($(main.west |- TrainA.west) + (0.2cm,0.0cm)$) -- (TrainA.west);
    \draw[-{Triangle},draw=black, line width=1pt] (base.east) -- ($(main.west |- base.east) + (0.2cm,0.0cm)$) -- ($(main.west |- TrainB.west) + (0.2cm,0.0cm)$) -- (TrainB.west);

    \node (Merge) [rounded corners, minimum height=\dheight, minimum width=1.6cm, fill=mypurple!40, anchor=west, align=center] at ($(TrainA.east)!0.5!(TrainB.east) + (0.6,0.0)$) {\bf Merge};

    \draw[-{Triangle},draw=black, line width=1pt] (TrainA.east) -- (Merge.north |- TrainA.east)  -- (Merge.north);
    \draw[-{Triangle},draw=black, line width=1pt] (TrainB.east) -- (Merge.south |- TrainB.east)  -- (Merge.south);

    \node (final) [rounded corners, minimum height=\dy, minimum width=\boxwidth, fill=gray!30, anchor=west, align=center] at ($(main.east |- Merge.east) + (0.6, 0.0)$) {\bf Final \\\bf Model};

    \draw[-{Triangle},draw=black, line width=1pt] (Merge.east) -- (final.west);

    \draw[-{Triangle},draw=black, line width=1pt] ($(main.east |- Merge.east) + (0.2cm,0.0cm)$) -- ($(main.north east) + (0.2cm,0.2cm)$) -- ($(main.north west) + (-0.2cm,0.2cm)$) -- ($(main.west |- base.east) + (-0.2cm,0.0cm)$);

    \node[anchor=west] at ($(main.east |- Merge.east)!0.5!(main.north east) + (0.2cm,0.4cm)$) {\bf$\times \frac{N-2}{2}$};

    
    \node (dataA) [minimum height=\dheight, minimum width=0.5*\dwidth, fill=mygreen!20, anchor=south east, align=center] at ($(main.south) + (0, 0.6)$) {};
    \node (dataA) [rounded corners,minimum height=\dheight, minimum width=\dwidth, fill=mygreen!20, anchor=east, align=center] at ($(dataA.east) + (0, 0.0)$) {$\mathcal{X}_i$};
    \node [minimum height=\dheight, minimum width=0.5*\dwidth, fill=mygreen!40, anchor=west, align=center] at ($(dataA.east) + (0, 0.0)$) {};
    \node (dataB) [rounded corners, minimum height=\dheight, minimum width=\dwidth, fill=mygreen!40, anchor=west, align=center] at ($(dataA.east) + (0, 0.0)$) {$\mathcal{X}_{i+1}$};

    \draw[-{Triangle},draw=black, line width=1pt] ($(main.south) + (0.0cm,-0.0cm)$) -- ($(main.south) + (0.0cm,0.2cm)$) -- ($(dataA.south) + (0.0cm, -0.4cm)$) -- ($(dataA.south) + (0.0cm,-0.0cm)$);
    \draw[-{Triangle},draw=black, line width=1pt] ($(main.south) + (0.0cm,-0.0cm)$) -- ($(main.south) + (0.0cm,0.2cm)$) -- ($(dataB.south) + (0.0cm, -0.4cm)$) -- ($(dataB.south) + (0.0cm,-0.0cm)$);

    \coordinate (midA) at ($(dataA.north)!0.35!(TrainB.south)$) {};
    \coordinate (midB) at ($(TrainA.north)!0.55!(TrainB.south)$) {};
    
    \draw[-{Triangle},draw=black, line width=1pt] (dataA.north) -- (dataA.north |- midA) -- (TrainB.south |- midA) -- (TrainB.south);

    \draw[draw=gray!10, line width=6pt] (dataB.north) -- (dataB.north |- midB) -- (TrainA.south |- midB) -- (TrainA.south);
    \draw[-{Triangle},draw=black, line width=1pt] (dataB.north) -- (dataB.north |- midB) -- (TrainA.south |- midB) -- (TrainA.south);


    \node (data2) [minimum height=\dheight, minimum width=\dwidth, fill=myblue!27, anchor=north east, align=center] at ($(main.south) + (0, -0.4)$) {$\mathcal{X}_2$};
    \node [minimum height=\dheight, minimum width=0.5*\dwidth, fill=myblue!15, anchor=east, align=center] at ($(data2.west) + (0, 0.0)$) {};
    \node (data1) [rounded corners, minimum height=\dheight, minimum width=\dwidth, fill=myblue!15, anchor=east, align=center] at ($(data2.west) + (0, 0.0)$) {$\mathcal{X}_1$};
    \node (data3) [minimum height=\dheight, minimum width=\dwidth, fill=myblue!39, anchor=west, align=center] at ($(data2.east) + (0, 0.0)$) {$\cdots$};
    \node [minimum height=\dheight, minimum width=0.5*\dwidth, fill=myblue!51, anchor=west, align=center] at ($(data3.east) + (0, 0.0)$) {};
    \node (data4) [rounded corners, minimum height=\dheight, minimum width=\dwidth, fill=myblue!51, anchor=west, align=center] at ($(data3.east) + (0, 0.0)$) {$\mathcal{X}_N$};

    \draw[-,draw=black, line width=1pt] ($(main.south) + (0.0cm,-0.0cm)$) -- ($(main.south) + (0.0cm,-0.2cm)$) -- ($(data1.north) + (0.0cm, 0.2cm)$) -- ($(data1.north) + (0.0cm,-0.0cm)$);
    \draw[-,draw=black, line width=1pt] ($(main.south) + (0.0cm,-0.0cm)$) -- ($(main.south) + (0.0cm,-0.2cm)$) -- ($(data2.north) + (0.0cm, 0.2cm)$) -- ($(data2.north) + (0.0cm,-0.0cm)$);
    \draw[-,draw=black, line width=1pt] ($(main.south) + (0.0cm,-0.0cm)$) -- ($(main.south) + (0.0cm,-0.2cm)$) -- ($(data3.north) + (0.0cm, 0.2cm)$) -- ($(data3.north) + (0.0cm,-0.0cm)$);
    \draw[-,draw=black, line width=1pt] ($(main.south) + (0.0cm,-0.0cm)$) -- ($(main.south) + (0.0cm,-0.2cm)$) -- ($(data4.north) + (0.0cm, 0.2cm)$) -- ($(data4.north) + (0.0cm,-0.0cm)$);

    \node[anchor=east] at ($(data1.west) + (0.0cm,0.0cm)$) {\bf Training Data};


\end{tikzpicture}
    }
    \caption{Illustration of \methodl (\method). We first split the training data into $N$ slices (blue \protect\boxb). We then iteratively finetune the current base model on two of these slices (green \protect\boxg) and merge the resulting models to obtain the base model for the next iteration (purple \protect\boxp). We repeat this until all $N$ data slices have been used.}
    \label{fig:method}
\end{figure}
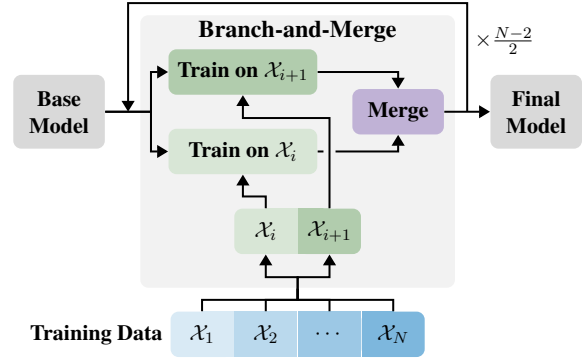

\paragraph{Experience Replay}
To mitigate such catastrophic forgetting, mixing in source \task data in the target \task training set, so-called experience replay, has proven effective for both continued pretraining \citep{simple} and instruction tuning  \citep{ScialomCM22,ZhangF0N23}.  
However, experience replay alone can not fully mitigate forgetting. Especially when the exact source data is unknown (e.g. for state-of-the-art language models), experience replay can only be implemented approximately, reducing its effectiveness and necessitating further regularization.

\paragraph{This Work: Mitigating Catastrophic Forgetting with Branch-and-Merge}
We build on ideas from continual learning and introduce \methodl (\method~-- illustrated in \cref{fig:method}), a novel method for adapting pretrained language models to new languages, underrepresented in their unknown training data, while minimizing the loss of previously learned capabilities.
Concretely, \method splits the training data into $N$ slices (blue \boxb in \cref{fig:method}), before iteratively training the current base model on $K$ (here two) such data slices in parallel (green \boxg) and finally merging them (purple \boxp) to obtain the initial model for the next iteration. This significantly reduces the total weight change and as a result, forgetting, while preserving most of the learning from the parallel training steps. In particular, while target language perplexity is slightly increased compared to standard continued training, the retained base model skills lead to higher downstream performance on target language tasks.

\paragraph{Results}
We apply \method to adapt \mistral[7B] \citep{Mistral7B} and \llama[3][8B] \citep{llama3modelcard} from predominantly English to an alphabet-sharing (German) and a non-alphabet-sharing (Bulgarian) language, considering both continued pretraining and instruction tuning.

We show that \method consistently improves benchmark performance in both the target and source \task compared to standard training, while not incurring additional computational or data costs. For example, when applied to instruction tuning, \method significantly improves performance, allowing our \llama[3][8B] \method-trained for Bulgarian to outperform \llama[3-8B][Instruct] not only in Bulgarian (by $10.9\%$) but also in English (by $1.3\%$) by inducing smaller magnitude but more efficient weight changes. In particular, we show that \method induces more favorable trade-offs between learning and forgetting than prior techniques such as reduced learning rates \citep{WinataXRWJ0KP23} and \lora \citep{biderman2024lora}.

\paragraph{Key Contributions} Our main contributions are:
\begin{itemize}
    \item We propose \methodl (\method), a training technique for language adaptation, improving learning while mitigating forgetting (\cref{sec:method}).
    \item We develop a high-quality data mix for approximate experience replay significantly improving language transfer (\cref{sec:data}).
    \item We conduct an extensive empirical investigation demonstrating the effectiveness of \method across two target languages (\cref{sec:eval}).
\end{itemize}

\section{Model Merging}
A wide range of model merging methods have been proposed \citep{MatenaR22,YadavTCRB23,StoicaBBHH23,YuYYHL23,WortsmanIGRLMNF22}. We experiment with \linear \citep{WortsmanIGRLMNF22}, \slerp \citep{goddard2024arcee,Shoemake85} and \modelstock \citep{JangYH24} merging, focusing on the first two, explained below. Let us consider the pretrained base model, $f_\theta$, parameterized by $\theta$ which was finetuned on two different datasets $\bc{X}_1$ and $\bc{X}_2$, yielding $f_{\theta_1}$ and $f_{\theta_2}$, respectively. We call the changes in weight due to this finetuning the task vectors $\tau_i = \theta_i - \theta$. To obtain a single model combining the learning from both datasets, we now merge these models. 

\linear model merging interpolates task vectors or equivalently parameterizations linearly so to obtain $\theta' \coloneqq \linear(\theta_1, \theta_2, c) = (1-c)\, \theta_1 + c\, \theta_2$.

\slerp first represents task vectors in polar coordinates before interpolating to obtain the new parameterization $\theta' = \tau' + \theta$
\begin{align*}
    \vartheta =& \arccos{\frac{\tau_1 \cdot \tau_2}{|\tau_1| \cdot |\tau_2|}}    \\
    \tau' =& \frac{\sin((1-c) \,\vartheta)}{\sin(\vartheta)} \tau_1 + \frac{\sin(c \, \vartheta)}{\sin(\vartheta)} \tau_2
\end{align*}
where $\vartheta$ is the angle between the two parameterizations and $c$ is the interpolation coefficient. By slight abuse of notation, we write $\slerp(\theta_1, \theta_2, c)$ for both the resulting parameters $\theta'$ and the corresponding model $f_{\theta'}$.

\section{\methodl for Mitigating Forgetting in Language Transfer}\label{sec:method}

To adapt a model $f_\theta$ pretrained on a typically unknown data distribution $\bc{X}_{\text{pre}}$ to a new task (language) without suffering from catastrophic forgetting, we propose the \methodl (\method) method, visualized in \cref{fig:method}. \method is based on first splitting the available training data into $N$ slices (blue \boxb in \cref{fig:method}), and then iteratively training $K$ models in parallel on one slice each (green \boxg) before merging the resulting models to obtain the base model for the next training iteration (purple \boxp).
We first provide the intuition behind \method before describing it in more detail.

\begin{figure}[t]
    \centering
    \resizebox{0.85\linewidth}{!}{
    \input{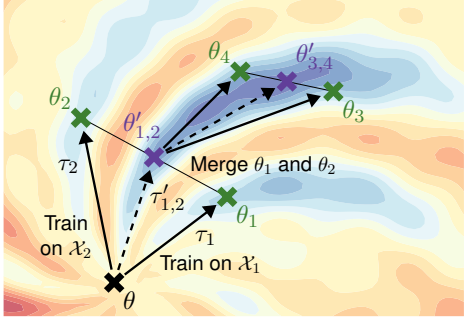}
    }
    \vspace{-3mm}
    \caption{Illustration of \method in the loss surface over parameter space. Both $\theta_1$ and $\theta_2$ land in poor local minima but their merge $\theta'_{1,2}$ lies in the valley of a better minimum. Training from there, $\theta_3$ and $\theta_4$ land at the boundary of that minimum due to noise in the training process and limited data. Their merge $\theta'_{3,4}$ cancels these errors and lies in the better minimum.}
    \label{fig:intuition}
    \vspace{-4mm}
\end{figure}

\paragraph{Intuition} There are two key ideas underlying \method. First, lower magnitude weight changes $\tau_i$, called task vectors, lead to less forgetting but also less learning. Second, the randomness in finetuning leads to task vectors $\tau_i = \tau^* + \epsilon_i$ with an unbiased error $\epsilon_i$ around the locally optimal task vector $\tau^*$ \citep{JangYH24}. We can thus reduce forgetting by reducing the task vector magnitude while offsetting the reduced learning by increasing task vector quality, i.e., reducing the error $\epsilon$. If this error is unbiased and empirically Gaussian $\epsilon_i \sim \bc{N}(0, \sigma^2)$ \citep{JangYH24}, merging, i.e., averaging, $K$ noisy task vectors to obtain $\tau' = \frac{1}{K} \sum_{i=1}^K \tau_i$ reduces the corresponding expected error magnitude with $||\epsilon'||_2 \propto \frac{1}{\sqrt{K}}$, as $\tau' \sim \bc{N}(\tau^*, \frac{1}{K}\sigma^2)$. At the same time, increasing $K$ in \method reduces the number of consecutive training iterations (as more data slices are used per iteration) and thus the expected total weight magnitude which in turn reduces both learning and forgetting. 
This allows \method to trade off learning and forgetting.

We visualize this in \cref{fig:intuition} for $K = 2$. There, the first two task vectors $\tau_1$ and $\tau_2$ land in the basins of poor local minima, with their merge $\theta'_{1,2}$ falling into the basin of a better minimum, highlighting the importance of \method's iterative merging approach. Training $\theta'_{1,2}$ on two more data slices yields noisy task vectors $\tau_3$ and $\tau_4$ at the edge of this loss basin with their merge $\theta'_{3,4}$ falling right in the middle. 
In contrast, simply reducing the learning rate can also reduce task vector magnitude but does not improve task vector quality. We note that the same intuitions apply to \slerp and \modelstock merging.

\paragraph{Implementation} In more detail, we first partition the training data $\bc{X}_{\text{train}}$ into $N$ not necessarily i.i.d. or equal-sized data slices $\bc{X}_i$. Then, we choose a parallelism factor $K$ ($K\!=\!2$ for most experiments and the visualizations in \cref{fig:method,fig:intuition}) and train our current base model $f_{\theta}$ independently on $K$ of these data slices yielding $f_{\theta_i}$ to $f_{\theta_{i+K-1}}$. We merge the resulting models to obtain the base model for the next iteration $f_{\theta'} = \merge(\{f_{\theta_j}\}_{j=i}^{i+K-1}, c)$. We typically choose the merging coefficient $c=0.5$ but note that we can easily perform a $1$-d line search over the resulting models. We then set the merged model $f_{\theta'}$ to be the base model $f_{\theta} \gets f_{\theta'}$ for the next training iteration and repeat this process until we have used all data slices. We formalize this approach in \cref{alg:bam}.

\begin{figure}[t]
\vspace{-3mm}
\begin{algorithm}[H]
    \caption{\methodl (\method)}
    \label{alg:bam}
    \begin{algorithmic}[1]
    \REQUIRE $K$: parallelism factor, $f_\theta$: base model, $\{\bc{X}_i\}_{i=1}^N$: data slices, $c$: merging coefficient
    \STATE $\Theta \gets \{\}$
    \FOR{$i \in [N]$}
    \STATE $f_{\theta_i} \gets \text{train}(f_\theta, \bc{X}_i)$
    \STATE $\Theta \gets \Theta \cup \{\theta_i\}$
    \IF{$i \! \mod \! K = 0 \, || \, i = N$}
    \STATE $\theta \gets \merge(\Theta,c)$
    \STATE $\Theta \gets \{\}$
    \ENDIF
    \ENDFOR
    \RETURN $f_\theta$: finetuned model
\end{algorithmic}
\end{algorithm}
\vspace{-8mm}
\end{figure}


\section{Data Mixtures for Mitigating Forgetting in Language Transfer}\label{sec:data}
Here we describe the data we use for continued pretraining of predominantly English base language models in order to adapt them to other languages. Outside of training methodology, we find in agreement with prior work that high-quality dataset mixtures are paramount for both effective language adaptation and reducing forgetting. We distinguish between experience replay of source language data and target language training data.

\subsection{Approximate Experience Replay of Source Domain Data} \label{sec:approx_replay_data}
While experience replay is crucial to alleviate forgetting \citep{RolnickASLW19,simple}, the training data of most state-of-the-art models remains undisclosed. We, therefore, rely on approximate experience replay, constructing our approximate source data based on prior work \citep{PenedoMHCACPAL23,together2023redpajama,TouvronLIMLLRGHARJGL23,olmo}. 

In more detail, we create a dataset consisting of OpenWebText \citep{openwebtext} - an open-source recreation of WebText \citep{gpt2}, English Wikipedia, GitHub repositories, and a range of instruction finetuning datasets with a total of $15.1$B unique tokens (see \cref{tab:data_source}). We repeat the smaller IFT datasets $4$ times to obtain an effective dataset size of 17.1B tokens. We note that while pretraining datasets commonly contain some instruction/response pairs, for example from Reddit, our experience replay mix most likely contains a higher portion of instruction data than the unknown source distribution.

\begin{table}[t]
\centering
\caption{Composition of the approximate experience replay dataset. We report the number of unique tokens, how often a dataset is repeated (Rep.), and the resulting sampling probability (Prob.).}
\vspace{-2mm}
\resizebox{\linewidth}{!}{
\begin{tabular}{llccc}
    \toprule
    Dataset & Domain & \#Tokens & Rep. & Prob. [\%]\\
    \midrule
    OpenWebText & Web & 8.5B & 1 & 49.8 \\
    Wikipedia-EN & Wiki & 4.6B & 1 & 26.9 \\
    GitHub repos & Code & 1.35B & 1 & 7.9 \\
    OpenHermes-2.5 & IFT & 357M & 4 & 8.4 \\
    SlimOrca & IFT & 197M & 4 & 4.6 \\
    MetaMathQA & IFT & 85M & 4 & 2.0 \\ 
    CodeInstructions & IFT & 20M & 4 & 0.47 \\ 
    \bottomrule
    \end{tabular}
}
\label{tab:data_source}
\vspace{-3mm}
\end{table}

\subsection{Minimal Experience Replay of Source Domain Data} \label{sec:data_min}
To explore the significance of high-quality data for experience replay, we contrast the aforementioned approximate experience replay with what we call \emph{minimal} experience replay. In minimal experience replay, we exclusively utilize samples from OpenWebText  \citep{openwebtext}, instead of a carefully curated data distribution. While minimal experience replay still incorporates source domain data during continuous pertaining, we anticipate it to cause a greater distribution shift than approximate experience replay. We chose the minimal experience replay to comprise roughly one-eighth of the training data ($5$B tokens for German and $10$B for Bulgarian).

\subsection{Constructing Target Language Data}
While designing an optimal training data mix is still an open research problem \citep{Xie0DDLLLL0Y23,TirumalaSAM23,ShenTMNLWTHVSX23}, some key components have been identified that we adhere to for our target domain data. In particular, it has been shown that a small portion of code can notably improve the resulting reasoning capabilities and should thus be included \citep{Liang23,MaLYZJWL23,fu2022gptroadmap}. Furthermore, the importance of reasoning and instruction-following capabilities for end tasks suggests that instruction data would benefit the continued pretraining data mix. This agrees well with  \citet{JiangSSRZNLYI24} suggesting a pre-instruction tuning phase to improve learning from new documents in continued pretraining.  We discuss the exact data mixes we use in \cref{sec:dataset}.

\paragraph{Bulgarian Training Data}
We adapt the RedPajama v2 pipeline \citep{together2023redpajama} for Bulgarian/Cyrillic to annotate $84$ Common Crawl \footnote{\url{https://commoncrawl.org/}} snapshots with a total of $30$T tokens. After aggressive quality filtering and near-deduplication, we obtain a dataset of $50$ to $80$B Bulgarian tokens, depending on tokenization. We augment this dataset using the Bulgarian split of publicly available multilingual high-quality datasets such as Wikipedia, Eur-lex \citep{baisa-etal-2016-european}, Europarl \citep{koehn-2005-europarl}, Parlamint \citep{parlamint}, books, and a selection of private datasets containing news articles, legal texts, and literature. We further include selected machine-translated instruction data. See \cref{tab:data_bg} for a full list of datasets. This yields a total of $77.7$B unique tokens (using the original \llama[3] tokenizer) which we boost to 82.1B tokens by repeating the smaller and particularly high-quality datasets between $2$ and $4$ times.

\paragraph{German Training Data}
German is significantly more abundant than Bulgarian in the quantity of text available from public datasets. We thus subsample roughly $10\%$ of the German subset from the curated web text dataset CulturaX \cite{culturax} equal to $41$B \llama[3] tokens and include three German IFT datasets. For more details, see \cref{tab:de_ift_data}. 

\begin{table}[t]
\centering
\caption{Composition of the Bulgarian target domain dataset. We report the number of unique tokens, how often a dataset is repeated (Rep.), and the resulting sampling probability (Prob.).}
\vspace{-2mm}
\resizebox{\linewidth}{!}{
\begin{tabular}{llccc}
    \toprule
    Dataset & Domain & \#Tokens & Rep. & Prob. [\%] \\
    \midrule
    RPv2-BG & Web & 70B & 1 & 85.3 \\
    Legal docs & Legal & 4.3B & 1 & 5.2 \\
    Books & Literature & 2.4B & 2 & 5.9 \\
    Eur-Lex-BG & Legal & 337M & 2 & 0.82 \\
    Wikipedia-BG & Wiki & 251M & 4 & 1.2 \\
    OrcaMath-BG & IFT & 100M & 4 & 0.49 \\
    Bulgarian Law & Legal & 58M & 4 & 0.28 \\
    Parlamint-BG & Transcripts & 52M & 3 & 0.19 \\
    Curlicat & Mixed & 40M & 2 & 0.10 \\
    SlimOrca-BG & IFT & 36M & 4 & 0.18 \\
    CodeInstructions-BG & IFT & 26M & 4 & 0.13 \\
    Europarl-BG & Transcripts & 24M & 3 & 0.09 \\
    MetaMath-BG & IFT & 15M & 4 & 0.07 \\
    Open-Platypus-BG & IFT & 13M & 4 & 0.06 \\
    \bottomrule
    \end{tabular}
}
\label{tab:data_bg}
\vspace{-3mm}
\end{table}

\section{Experimental Setup}
We now describe the experimental setup used to validate \method's effectiveness for \task adaptation. In particular, we discuss the target languages (Bulgarian and German), evaluation benchmarks (\cref{sec:eval_data}), training data (\cref{sec:dataset}), and training setup (\cref{sec:training_setup}). We experiment with both continued pretraining of base models and instruction tuning of the resulting models.

\subsection{Target Languages and Benchmarks}\label{sec:eval_data}
To evaluate the effectiveness of \method we conduct experiments on the transfer from general purpose, predominantly English models, to an alphabet-sharing (German) and a non-alphabet-sharing (Bulgarian) language, evaluating the resulting models on both the target and source languages.

While there is a large and growing number of high-quality datasets for evaluating LLMs in  English and to a lesser extent German, these are much sparser for low-resource languages such as Bulgarian. We therefore first provide a brief overview of the English and German benchmarks we use before discussing the construction of a holistic evaluation suite for Bulgarian.

\paragraph{English Benchmarks}
We consider the following domains and benchmarks in English: \emph{commonsense reasoning} (HellaSwag \citep{hellaswag}, Winogrande \citep{winogrande}, ARC-Easy, ARC-Challenge \citep{arc}), \emph{multitask capabilities} (MMLU \citep{mmlu}), \emph{math} (GSM8K \citep{gsm8k}, MathQA \citep{mathqa}), and \emph{reading comprehension} (Belebele English \citep{belebele}, TriviaQA \citep{triviaqa}). We provide a detailed description of these benchmarks in \cref{sec:app_benchmarks}.
\paragraph{German Benchmarks}
We use the German benchmarks available in the Language Model Evaluation Harness \citep{eval-harness}. 
Some of these benchmarks are translated from English using GPT 3.5 \citep{german_benchmarks} (TruthfulQA-DE, HellaSwag-DE, MMLU-DE, ARC-DE). We further consider human curated or translated benchmarks for \emph{math} (MGSM-DE \citep{mgsm}), \emph{paraphrasing} (PAWS-X \cite{pawsx}) and \emph{reading comprehension} (BeleBele German \citep{belebele}). A detailed description of these benchmarks can be found in \cref{sec:app_benchmarks}.

\paragraph{Bulgarian Benchmarks}
As the number of publicly available Bulgarian benchmarks is limited, we translate all of the above English benchmarks using a combination of machine translation and over $600$ hours of professional translators' work. We denote the translated benchmarks by appending `-BG' to their name and make them publicly available. In addition, we use the following Bulgarian benchmarks: 
\emph{natural language inference} (XNLI \citep{xnli}) and \emph{high-school exams} (EXAMS \citep{exams}, MON-Tests). From these, XNLI was constructed through a professional translation of English examples by ~\citeauthor{xnli} and the other two are natively Bulgarian.  We provide more details on the construction of the translation process and the novel MON-Tests benchmark in \cref{sec:translation}.

\paragraph{Evaluation Metrics}
We aim to measure both learning, i.e., language adaptation, and forgetting. To this end, we consider benchmark scores and perplexity in the source and target language. Since our approximate experience replay data contains instruction tuning examples which can lead to improved English benchmark scores compared to the base model, we focus on held-out English document perplexity as a measure of forgetting. We use both benchmark performance (normalized accuracy) and held-out document perplexity as a measure of learning in the target language (see \cref{sec:app_data} for more details). 

For both English and Bulgarian, we evaluate MMLU, TriviaQA, and EXAMS in a 5-shot, GSM8K in an 8-shot, and all other benchmarks in a zero-shot setting. All German benchmarks are run in a 5-shot setting.

\subsection{Training Data}\label{sec:dataset}
Below, we discuss the training data used for language adaptation.

\paragraph{Continued Pretraining Data}

For German, we subsample the training data including the approximate ($17$B tokens) and minimal experience replay ($5$B tokens) to $50$B and $40$B tokens respectively and divide it into $N=4$ i.i.d.\ slices of $12.5$B and $10$B tokens each. 
For Bulgarian, we split the full $82$B tokens of Bulgarian data plus $17$B tokens of approximate or $10$B tokens of minimal experience replay into $N=8$ slices either i.i.d.\ or via a curriculum where the even-numbered slices contain significantly more experience replay data than the odd ones (see \cref{tab:curriculum} in \cref{sec:app_data}).

\paragraph{Instruction Finetuning Data}
We investigate the effectiveness of \method for instruction finetuning after continued pretraining. We collect $928K$ samples of English finetuning data and mix it with German or Bulgarian data. For Bulgarian, we generate $78K$ samples by using a mix of machine translation and professional translators to translate English samples to Bulgarian. For German, we use a mix of available, translated German IFT datasets. Please see \cref{tab:en_ift_data,tab:bg_ift_data,tab:de_ift_data}, as well as \cref{sec:app_data} for details on the resulting dataset.

\subsection{Training Setup} \label{sec:training_setup}
\paragraph{Base models}
We chose \mistral[7B] \citep{Mistral7B} and \llama[3][8B] \citep{llama3modelcard} as base models due to their exceptional performance for their size and permissive licenses. 

\paragraph{Details} We implement \method in PyTorch \citep{PaszkeGMLBCKLGA19} using HuggingFace's transformers library \citep{WolfDSCD19} and DeepSpeed 
\citep{deepspeed10.1145/3394486.3406703,zero9355301}.
We train each model on 64 NVIDIA H100s. Based on prior work and initial experiments, we find that $10^{-5}$ is the best maximum learning rate for continued pretraining on the models that we are using together with a batch size of $512$ for continued pretraining and $256$ for supervised finetuning. We use cosine decay to $0.1 \cdot \text{max\_lr}$ with $\max(100,0.01\cdot\text{total\_steps})$ linear warmup.

\begin{table*}[t]
\centering
\caption{Effect of \method with $N=8$ and $K=2$ for the language transfer to Bulgarian. We report normalized accuracies and their averages with full results on English benchmarks deferred to \cref{tab:cpt_eng}.}
\label{tab:cpt_bg}
\vspace{-3mm}
\resizebox{\linewidth}{!}{
\begin{tabular}{lll cccccccccccc cccc}
\hline
    \toprule
    \multirow{2.5}{*}{Model} & \multirow{2.5}{*}{CPT} & \multirow{2.5}{*}{IFT} & \multicolumn{12}{c}{Bulgarian Benchmarks}& \multirow{2.5}{*}{Avg BG} &\multirow{2.5}{*}{Avg EN}
    &\multirow{2.5}{*}{BG NLL}
    &\multirow{2.5}{*}{EN NLL}\\
    \cmidrule(lr){4-15}

    && & WG & HS & ARC-c & ARC-e & MMLU & MathQA & GSM8K & TrQA & MON & Bele & XNLI & EXAMS &  & \\
    \midrule
    \multirow{6.5}{*}{L-8B} & \multicolumn{2}{c}{\llama[3][8B (Base)]} & $61.48$ & $52.19$ & $34.89$ & $53.32$ & $50.81$ & $35.37$ & $36.99$ & $27.19$& $40.91$ & $45.77$ & $45.46$ & $45.75$  & $44.18$ &  $63.85$ & $1.695$ & $\textbf{2.042}$\\
    \cmidrule(lr){2-2}
    & Standard &  -&                 $67.40$ & $\textbf{66.48}$ & $42.32$ & $62.37$ & $53.02$ & $\textbf{38.65}$ & $54.81$ &  $\textbf{38.69}$& $46.29$ & $\textbf{62.11}$ & $\textbf{48.75}$ & $\textbf{56.43}$ & $53.11$ & $64.84$ & $\textbf{1.018}$ & $2.138$\\

    & Half LR  & - & $68.67$&$65.97$&$40.36$&$62.54$&
$53.71$&$37.89$&$56.79$&$37.96$&
$46.71$&$60.89$&$47.51$&$52.60$&
$52.63$&$65.06$&$1.055$&$2.098$\\
    &\method  & -&             $\textbf{69.92}$ & $66.14$ & $\textbf{42.66}$ & $\textbf{63.17}$ & $\textbf{54.29}$ & $37.48$ & $\textbf{59.43}$ & $38.53$& $\textbf{46.92}$ & $59.66$ & $48.03$ & $52.60$   & $\textbf{53.40}$& $\textbf{66.24}$&
    $1.061$&
    $2.097$\\
    \cmidrule(lr){2-3}
    & \multicolumn{2}{c}{\llama[3][8B-Instruct]} &$58.64$&$48.91$&$34.47$&$50.88$&$49.71$&$33.63$&$55.80$&$26.79$&$40.65$&$64.00$&$44.74$&$45.48$&$46.14$&$68.72$& $1.950$ &$2.307$\\
    \cmidrule(lr){3-3}

    & \method  & Standard & $68.67$&$66.75$&$47.95$&$\textbf{70.24}$&
$52.54$&$38.73$&$63.84$&$31.70$&
$48.60$&$\textbf{80.44}$&$50.92$&$51.51$&
$55.99$&$67.69$& $1.208$ &$2.290$ \\

    & \method  & \method & $\textbf{68.98}$&$\textbf{68.01}$&$\textbf{49.57}$&$69.07$&
$\textbf{54.04}$&$\textbf{38.56}$&$\textbf{65.05}$&$\textbf{36.17}$&
$\textbf{49.94}$&$79.22$&$\textbf{51.45}$&$\textbf{53.42}$&
$\textbf{56.96}$&$\textbf{69.97}$& $\textbf{1.148}$ &$\textbf{2.193}$  \\
    \cmidrule(lr){1-3}
    \multirow{4}{*}{M-7B} & \multicolumn{2}{c}{\mistral[7B (Base)]}
    &$61.48$&$53.63$&$37.54$&$55.93$&$49.37$&$31.36$&$29.04$&$29.32$&$42.15$&$39.67$&$42.77$&$44.93$&$43.10$&$59.81$&$1.525$&$\textbf{1.883}$\\
    \cmidrule(lr){2-2}
    & Standard & - 
&$68.19$&$67.20$&$41.13$&$57.95$&$52.41$&$33.87$&$\textbf{65.73}$&$42.08$&$46.85$&$51.44$&$45.10$&$53.97$&$52.16$&$62.03$&$\textbf{1.408}$&$1.967$\\
    & \method i.i.d. & - &
$69.77$&$\textbf{67.66}$&$41.04$&$60.01$&$\textbf{53.66}$&$34.61$&$58.23$&$41.78$&$45.60$&
$52.67$&$47.11$&$53.15$&$52.11$&$\textbf{63.72}$&$1.411$&$1.951$\\
    & \method & - &
$\textbf{70.24}$&$67.45$&$\textbf{43.26}$&$\textbf{61.62}$&
$52.63$&$\textbf{35.58}$&$59.97$&$\textbf{42.24}$&
$\textbf{46.98}$&$\textbf{52.78}$&$\textbf{48.23}$&$\textbf{54.79}$&$\textbf{52.98}$&$63.53$&$1.426$&$1.950$\\

    \bottomrule
\hline
\end{tabular}
}
\end{table*}

\begin{table*}[t]
\centering
\caption{Effect of \method with $N=4$ and $K=2$ for the language transfer of \llama[3][8B] to German. We report normalized accuracies and their averages with full results on English benchmarks deferred to \cref{tab:cpt_de_eng}.
  }
  \label{tab:cpt_de}
  \vspace{-3mm}
\resizebox{0.9\linewidth}{!}{
\begin{tabular}{llccccccccc}
    \toprule
    \multirow{2.5}{*}{CPT} & \multirow{2.5}{*}{IFT} & \multicolumn{7}{c}{German Benchmarks}& \multirow{2.5}{*}{Avg DE} &\multirow{2.5}{*}{Avg EN}\\
    \cmidrule(lr){3-9}

    && ARC-c & HellaSwag & MMLU & TruthfulQA & MGSM-DE & PAWS-DE & BeleBele  & \\
\midrule
\llama[3][8B (Base)] & - & $46.62$ & $62.03$ & $55.18$  & $46.51$ & $42.00$ & $36.15$ &  $81.22$ & $52.82$ & $63.85$\\
    \cmidrule(lr){1-1}
Standard min. replay& - & $47.98$ & $\textbf{66.49}$ & $55.23$  & $46.87$ & $41.20$ &  $37.80$ & $79.00$ & $53.51$ & $60.79$\\
\method min. replay& - & $47.21$ & $65.78$ & $55.62$  & $47.25$ & $44.40$ & $\textbf{39.80}$ &  $79.44$ & $54.22$ & $61.75$ \\

\method appx. replay& - & $\textbf{51.92}$ & $65.97$ & $\textbf{55.73}$  & $\textbf{54.33}$ & $\textbf{58.80}$ & $35.35$ &  $\textbf{81.67}$ & $\textbf{57.68}$ & $\textbf{65.79}$ \\
    \cmidrule(lr){1-1}

\method appx. replay& Standard & $\textbf{53.12}$ & $65.51$ & $54.60$  & $\textbf{55.20}$ & $\textbf{66.00}$ & $39.75$ &  $\textbf{86.44}$ & $60.09$ & $67.90$ \\

\method appx. replay& \method & $52.95$ & $\textbf{67.53}$ & $\textbf{55.80}$  & $54.28$ & $65.60$ & $\textbf{40.40}$ &  $85.89$ & $\textbf{60.35}$ & $\textbf{70.14}$ \\

\bottomrule
\end{tabular}
}
\vspace{-3mm}
\end{table*}

\section{Experimental Evaluation}\label{sec:eval}
We now evaluate \method empirically for both continued pretraining (CPT) and instruction finetuning (IFT) before conducting extensive ablations and providing further results in \cref{sec:extended_eval}.

\subsection{\method for Continued Pretraining}
\paragraph{Bulgarian CPT} We use our data mix of Bulgarian data and English experience replay to adapt both \llama[3][8B] and \mistral[7B] to Bulgarian, comparing standard CPT and \method in \cref{tab:cpt_bg}. We first demonstrate on \mistral that \method with i.i.d. data slices matches standard CPT in Bulgarian ($0.05\%$ average score difference) while reducing forgetting significantly ($20\%$ less English NLL increase), achieving a $1.7\%$ higher average score on English benchmarks and even outperforming the base model. Using our curriculum slices (only called \method), we outperform standard CPT in 11 out of 12 Bulgarian benchmarks while retaining the reduced forgetting. Similarly, \method achieves both a slightly higher average Bulgarian  ($0.3\%$ better) and a notably higher English score ($1.4\%$ better) for \llama[3]. We observed consistently across all of these settings that while standard CPT achieves a lower negative log-likelihood (NLL) in Bulgarian, indicating it fits the Bulgarian language modeling task better, the increased forgetting of base model capabilities (higher English NLL) leads to worse or equal benchmark performance.

\paragraph{German CPT} We observe very similar trends adapting \llama[3][8B] to German (see \cref{tab:cpt_de}) with \method outperforming standard CPT both in terms of German ($0.7\%$) and English ($1.0\%$) scores with minimal experience replay. Using our approximate experience replay and injecting German IFT data, further improves performance ($3.5\%$ in German and $4\%$ in English), surpassing the base \llama[3][8B] model now in both German and English benchmarks.

\subsection{\method for Instruction Fine-Tuning}
We investigate the effectiveness of \method for instruction finetuning, reporting results in \cref{tab:cpt_bg,tab:cpt_de}. We observe that \method slightly improves learning of both German and Bulgarian, while significantly reducing forgetting. Considering a wider range of settings in \cref{tab:ift_bg}, we observe that \method with $N=K=2$ and i.i.d. data slices not only strictly outperforms standard IFT on the combined data (IFT full) and an equal mix of Bulgarian and English data (IFT 50-50), but also IFT on just English data (IFT EN). Slicing the data by language (\method BG | EN) results in even greater improvements and outperforms \llama[3-8B][Instruct] \citep{llama3instruct} in both Bulgarian ($10.8\%$) and English ($1.3\%$). We hypothesize that merging the task vectors of IFT on multiple languages removes a lot of language-specific errors leaving a higher quality instruction following task vector.

\begin{table}[t]
    \centering
    \caption{\method for Bulgarian instruction tuning of our \method trained \llama[3][8B].}
    \vspace{-3mm}
    \label{tab:ift_bg}
    \resizebox{0.6\linewidth}{!}{
    \begin{tabular}{lcc}
         \toprule
         Method & Avg BG & Avg EN \\
         \midrule
         Base (\method trained) & 53.16 & 66.18 \\
         \cmidrule(lr){1-1}
         IFT full & 55.99 &  67.69 \\
         IFT 50-50 & 55.01 & 67.55  \\
         IFT EN & 54.72 & 67.76 \\
         IFT BG & 54.16 & 66.96 \\
         \method i.i.d.  & 56.45 & 68.65  \\
         \method BG | EN& \textbf{56.96} & \textbf{69.97} \\
         \cmidrule(lr){1-1}
         \llama[3][Instruct] & 46.14 & 68.72  \\
         \bottomrule
    \end{tabular}
    }
    \vspace{-3mm}
\end{table}

\subsection{Ablations}
Below, we investigate various components and design decisions underlying \method using the domain adaptation to Bulgarian. 

\begin{figure}[h]
    \centering
    \includegraphics[width=1.0\linewidth]{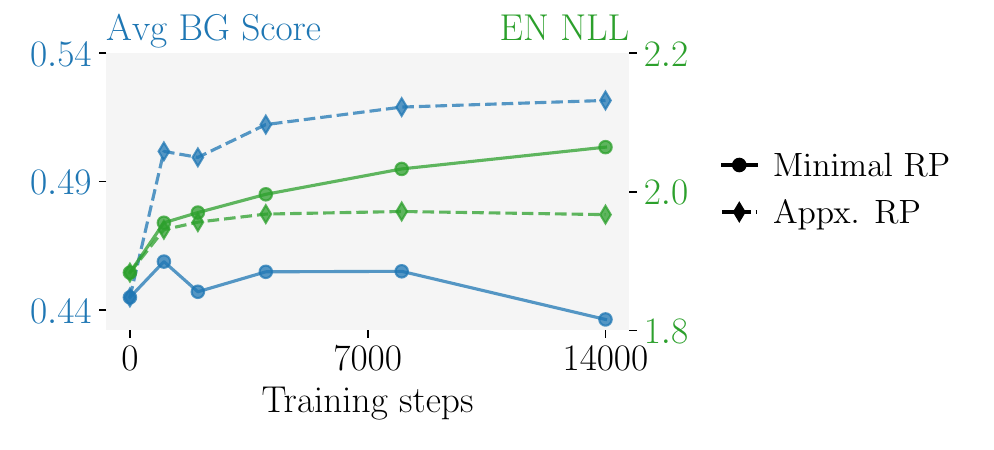}
    \vspace{-9mm}
    \caption{Comparing minimal and our approximate experience replay on \mistral with respect to average Bulgarian benchmark scores ($\uparrow$) and Negative Log-Likelihood (NLL) on the English validation set ($\downarrow$).}
    \vspace{-5mm}
    \label{fig:replay}
\end{figure}

\begin{table}[t]
    \centering
    \caption{Effect of approximate and minimal replay on source and target domain performanc. \method is with i.i.d. data slices.}
    \vspace{-1mm}
    \label{tab:bg_de_replay}
    \resizebox{0.9\linewidth}{!}{
    \begin{tabular}{llllccc}
         \toprule
         Model & Language & CPT & Replay & Avg BG & Avg DE & Avg EN  \\
         \midrule
         \multirow{3.5}{*}{\llama[3][8B]} &  \multirow{3.5}{*}{DE}& \multirow{2}{*}{min} & Standard  & - & 53.51  & 60.80 \\
         &  &  & \method  & - & 54.22 & 61.75 \\
         \cmidrule(lr){3-3}
         &  & appx & \method  & - & \textbf{57.68} & \textbf{65.79} \\
         \cmidrule(lr){1-4}
         \multirow{4.5}{*}{\mistral[7B]} &  \multirow{4.5}{*}{BG}& \multirow{2}{*}{min} & Standard  & 43.71 & -  & 51.44 \\
         &  &  & \method & 46.23 & - & 54.52 \\
         \cmidrule(lr){3-3}
         &  & \multirow{2}{*}{appx} & Standard  & \textbf{52.16} & - & 62.03 \\
         &  &  & \method & 52.11 & - & \textbf{63.72} \\
         
         \bottomrule
    \end{tabular}
    }
    \vspace{-3mm}
\end{table}

\paragraph{Approximate Experience Replay}
We compare our approximate experience replay, described in \cref{sec:approx_replay_data}, to minimal experience replay, described in \cref{sec:data_min}, for continued pretraining in \cref{fig:replay}. We observe that using minimal replay (solid lines in \cref{fig:replay}), target language performance (Avg BG Score -- blue) first increases before dropping off as capabilities of the base model are forgotten (increasing negative log-likelihood -- green). In contrast, using our approximate experience replay (dashed line), we see a much stronger increase in target domain performance and reduced forgetting of the source domain. 
We confirm these findings in German (see \cref{tab:bg_de_replay}) and thus use approximate experience replay for all other experiments.

\paragraph{\method and Experience Replay}
We compare the effectiveness of \method in the presence of minimal and approximate experience replay in \cref{tab:bg_de_replay} on Bulgarian, German and English benchmarks. We observe that \method is even more effective in the minimal replay setting, where the larger distribution shift induces more forgetting. There, \method can, e.g., improve the performance in Bulgarian and English by $2.5\%$ and $2.9\%$, respectively, compared to $0.0\%$ and $1.7\%$, respectively, in the approximate replay setting.

\begin{figure}[h]
    \centering
    \includegraphics[width=1.0\linewidth]{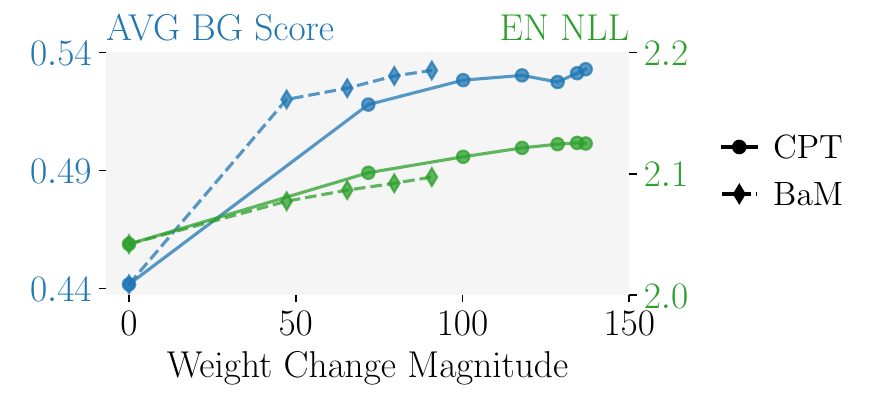}
    \vspace{-9mm}
    \caption{Average Bulgarian benchmark score ($\uparrow$) and English NLL ($\downarrow$) over L2 norm of weight change depending on training method for \llama[3]}
    \label{fig:weight_change_efficiency}
    \vspace{-5mm}
\end{figure}

\begin{figure}[h]
    \centering
    \includegraphics[width=1.0\linewidth]{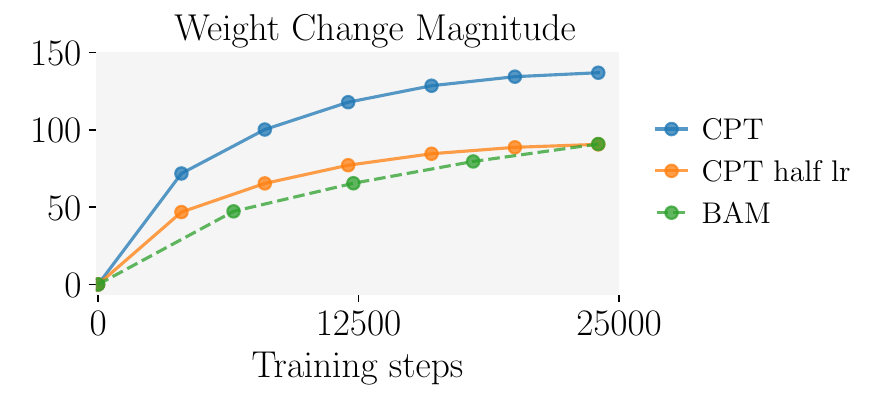}
    \vspace{-9mm}
    \caption{L2 norm of weight change depending on training method for \llama[3]}
    \label{fig:weight_change_magnitude}
    \vspace{-2mm}
\end{figure}
\paragraph{Forgetting and Weight Change Magnitude}
We plot the average BG score as a measure of learning and English NLL as a measure of forgetting over weight change magnitude in \cref{fig:weight_change_efficiency}. We observe that both forgetting and learning strongly correlated with weight change magnitude and that \method is more efficient, i.e., yields more learning and less forgetting at the same weight change, confirming our intuition discussed in \cref{sec:method}.

Comparing \method to standard CPT with a halved learning rate, we observe almost identical weight change magnitudes (see \cref{fig:weight_change_magnitude}) corresponding to $66\%$ of the standard CPT weight change. While the reduced learning rate CPT also reduces forgetting (although slightly less than \method), it comes at the cost of severely reduced learning (see \cref{tab:cpt_bg}). We observe a similar but stronger effect for \lora which only shows minimal learning (see \cref{tab:lora}).

\begin{figure}[h]
    \centering
    \includegraphics[width=0.75\linewidth]{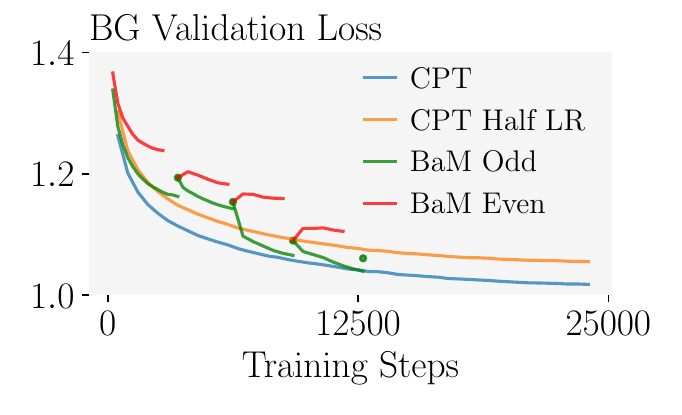}
    \vspace{-4mm}
    \caption{Bulgarian validation loss over training steps for \llama[3] depending on training method. \method Odd (green) is trained on more Bulgarian and \method Even (red) on more approximate experience replay. We show their merges as green dots.}
    \label{fig:training_dynamics}
    \vspace{-5mm}
\end{figure}
\paragraph{Training Dynamics with \method}
We compare the training dynamics of \method and standard CPT at full and half learning rate in \cref{fig:training_dynamics}. We observe that training on data slices with larger portions of experience replay (even -- red) cannot decrease Bulgarian validation loss further after a short period. However, after a merge, training on the Bulgarian-focused slices (odd -- green) converges significantly faster than for CPT at a similar validation loss, highlighting the potential of merging to escape local minima or flatter portions of the loss landscape.

\begin{table}[h]
    \centering
    \caption{Effect of of slice count $N$, parallelism factor $K$, and merging method on continued pertaining (CPT) of \llama[3] on a reduced Bulgarian dataset.}
    \vspace{-3mm}
    \label{tab:parallelism}
    \resizebox{0.95\linewidth}{!}{
    \begin{tabular}{lcccccc}
         \toprule
         Merging Method & N & K & Avg BG & Avg EN & BG NLL & EN NLL \\
         \midrule
          \multirow{1}{*}{-}  & \multicolumn{2}{c}{base} & $44.18$ & $63.85$ & $1.695$ & $2.042$ \\
         \cmidrule(lr){1-1}
          - & 1 & 1 & $51.76$ & $66.33$ & $\mbf{1.136}$ & $2.093$\\
         \cmidrule(lr){1-1}
         \multirow{3}{*}{\slerp} & 2 & 2 & $\mbf{52.01}$ & $\mbf{67.00}$ & $1.194$ & $2.077$\\
         & 4 & 2 & $51.88$ & $66.80$ & $1.186$ & $2.078$\\
         & 4 & 4 & $51.25$ & $66.76$ & $1.233$ & $\mbf{2.068}$\\
         \cmidrule(lr){1-1}
         \linear & 4 & 2 & $51.98$ & $66.65$ & $1.186$ & $2.078$\\
         \modelstock & 4 & 2 & $51.54$ & $66.98$ & $1.201$ & $2.069$\\
         \bottomrule
    \end{tabular}
    }
    \vspace{-3mm}
\end{table}
\paragraph{Effect of the Parallelism Factor}
We investigate the effect of the parallelism factor $K$ on dataset of $26$ B tokens, obtained by combining the first two data slices $\bc{X}_1$ and $\bc{X}_2$,  reporting results in \cref{tab:parallelism} where all settings use the same data and compute. 
We observe that training on all data jointly ($N=1, K=1$) reduces Bulgarian NLL the most but at the cost of increased forgetting (highest English NLL) leading to worse benchmark performance than \method. Comparing \method hyperparameters, we observe that increasing $K$ reduces both learning and forgetting as we reduce weight change magnitude and improve task vector quality ($N=4, K=4$). The best trade-off leading to the highest English and Bulgarian scores is attained with a parallelism factor of $K=2$ and data slices of roughly 10B tokens ($N=2, K=2$). We thus choose these settings for all other experiments leading to $N \in \{4,8\}$ for the full data.

\paragraph{Effect of the Merging Method}
We compare \slerp, \linear, and \modelstock merging in \cref{tab:parallelism} and observe that \slerp and \linear merging achieve almost identical results, with \modelstock reducing forgetting at the cost of reduced learning. As \slerp merging achieves slightly better scores, we use it for all other experiments.

\section{Related work}

\paragraph{Catastrophic Forgetting}
Neural networks trained on a specific task are known to catastrophically forget the previous task when adapted to a new one \citep{french1999catastrophic,GoodfellowMDCB13,KemkerMAHK18}.
While this becomes less pronounced as model and pertaining data size grow \citep{RamaseshLD22}, it remains a severe issue even for modern LLMs \citep{ZhaiTLCQLM23,ShiXWQWWW24,LiL24,GogoulouLBN23}.

\paragraph{Mitigating Catastrophic Forgetting} 
As LLMs are frequently finetuned or continually pretrained on new tasks, mitigating catastrophic forgetting has become essential and a wide range of methods has been proposed. \citet{LeeCK20} suggest to randomly reset weights to their pretrained state. \citet{biderman2024lora} show that \lora reduces forgetting at the cost of reduced learning. \citet{HuangCWYLSYS24} suggest experience replay with synthetic and \citet{simple} with original source domain samples. \citet{WinataXRWJ0KP23} propose to exponentially reduce the learning rate when learning new tasks. Similar to us, \citet{lin2024mitigating} suggest to linearly merge the adapted with the original model using block-wise parameters, focusing on alignment tuning instead of language transfer.

\paragraph{Model Merging}
Model merging was originally proposed in federated learning \citep{McMahanMRHA17} to lower communication costs, and was successfully deployed \citep{StoicaBBHH23,MatenaR22}. 
As a way to combine multiple models without training, it has recently gained popularity in the LLM community \citep{goddard2024arcee}. Popular methods include \linear or \modelar \citep{IlharcoRWSHF23} which perform linear interpolation of task vectors, its extension \modelbread \citep{DavariB23} which discards large weight changes,  \ties \citep{YadavTCRB23} which uses heuristics favoring large weight changes, \dares \citep{YuYYHL23} which randomly drops weight changes before merging, \modelstock \citep{JangYH24} which merge weights layer-wise, to in expectation, minimize distance to the center of the task vector distribution, and \slerp \citep{Shoemake85} which averages weights in polar coordinates.

Multiple works have shown that merging during continued pretraining or finetuning, especially on non-IID data, can match or improve the performance of compound training. 
\citet{lofi} average models finetuned without any communication.
\citet{btm} propose a scheme for iteratively branching and merging models during training, however, they assume the full data distribution is available for pertaining and focus on building ensembles rather than a single model.
\cold \citep{cold} is methodologically most similar to our work but focuses on training a base model which can then be easily adapted to a new task, rather than this adaptation itself. This objective is shared by \citet{choshen2022fusing} which only consider a single iteration of merging.

\section{Conclusion}
We proposed \methodl (\method) training to mitigate forgetting while boosting learning in language transfer by generating lower magnitude but higher quality weight changes. We showed that combining \method with an effective approximate experience replay data mix significantly reduces forgetting. Finally, we demonstrated that our approach can benefit both continuous pertaining and instruction tuning in both alphabet-sharing (German) and non-sharing (Bulgarian) languages. For instance, we outperform \llama[3][8B-Instruct] with the same base model in both source (English, $1.3\%$) and target (Bulgarian, $10.8\%$) languages.


\section{Limitations}

Our study focuses on language transfer to two languages with different characteristics and considers two models of up to 8 billion parameters. However, to establish the general applicability of our approach, potentially even to general domain adaptation, further experiments across a broader set of languages and tasks as well as model architectures will be necessary.

We considered specific data mixes for the continued pretraining in both considered languages which we observe to yield good performance --- it is possible that the success of \methodl depends on the composition of these datasets. While infeasible when adapting state-of-the-art pretrained models with unknown training set distribution, an evaluation of our method with exact experience replay would provide further understanding of its performance relative to the state-of-the-art in continuous learning, including joint training on all data.

While we consider a broad range of up to 12 benchmarks per language, they are still limited in their domain coverage. As \method does not outperform standard training across all benchmarks, this benchmark composition can affect the resulting conclusions.

While we originally optimized hyperparameters for standard training and carried them over to \method, it is possible, although unlikely, that a more extensive hyperparameter search would benefit standard training more than \method.

\section{Ethical Considerations}
We believe our work empowers practitioners to more efficiently adapt strong pretrained models to other potentially low-resource languages, thus contributing to the democratization of large language models. However, such models can of course also be abused and in particular if our approach generalizes beyond language to general domain adaptation, by malicious practitioners who could more efficiently adapt the models for nefarious tasks.

\section*{Acknowledgments}
This research was partially funded by the Ministry of Education and Science of Bulgaria (support for INSAIT, part of the Bulgarian National Roadmap for Research Infrastructure).

This work has been done as part of the EU grant ELSA (European Lighthouse on Secure and Safe AI, grant agreement no. 101070617). Views and opinions expressed are however those of the authors only and do not necessarily reflect those of the European Union or European Commission. Neither the European Union nor the European Commission can be held responsible for them.

The work has received funding from the Swiss State Secretariat for Education, Research and Innovation (SERI).

We would like to thank Dr. Maurice Weber for helping adapt the RPv2 pipeline to Bulgarian and create the Bulgarian dataset. We would also like to thank Hristo Venev for his help with system related issues.


\clearpage

\bibliography{references}

\begin{thebibliography}{95}
\providecommand{\natexlab}[1]{#1}

\bibitem[{AI@Meta(2024{\natexlab{a}})}]{llama3instruct}
AI@Meta. 2024{\natexlab{a}}.
\newblock \href {https://huggingface.co/meta-llama/Meta-Llama-3-8B-Instruct} {Llama 3 instruct details}.

\bibitem[{AI@Meta(2024{\natexlab{b}})}]{llama3modelcard}
AI@Meta. 2024{\natexlab{b}}.
\newblock \href {https://github.com/meta-llama/llama3/blob/main/MODEL_CARD.md} {Llama 3 model card}.

\bibitem[{Amini et~al.(2019)Amini, Gabriel, Lin, Koncel-Kedziorski, Choi, and Hajishirzi}]{mathqa}
Aida Amini, Saadia Gabriel, Shanchuan Lin, Rik Koncel-Kedziorski, Yejin Choi, and Hannaneh Hajishirzi. 2019.
\newblock \href {https://doi.org/10.18653/v1/N19-1245} {{M}ath{QA}: Towards interpretable math word problem solving with operation-based formalisms}.
\newblock In \emph{Proceedings of the 2019 Conference of the North {A}merican Chapter of the Association for Computational Linguistics: Human Language Technologies, Volume 1 (Long and Short Papers)}, pages 2357--2367, Minneapolis, Minnesota. Association for Computational Linguistics.

\bibitem[{Baisa et~al.(2016)Baisa, Michelfeit, Medve{\v{d}}, and Jakub{\'\i}{\v{c}}ek}]{baisa-etal-2016-european}
V{\'\i}t Baisa, Jan Michelfeit, Marek Medve{\v{d}}, and Milo{\v{s}} Jakub{\'\i}{\v{c}}ek. 2016.
\newblock \href {https://aclanthology.org/L16-1445} {{E}uropean {U}nion language resources in {S}ketch {E}ngine}.
\newblock In \emph{Proceedings of the Tenth International Conference on Language Resources and Evaluation ({LREC}'16)}, pages 2799--2803, Portoro{\v{z}}, Slovenia. European Language Resources Association (ELRA).

\bibitem[{Bandarkar et~al.(2023)Bandarkar, Liang, Muller, Artetxe, Shukla, Husa, Goyal, Krishnan, Zettlemoyer, and Khabsa}]{belebele}
Lucas Bandarkar, Davis Liang, Benjamin Muller, Mikel Artetxe, Satya~Narayan Shukla, Donald Husa, Naman Goyal, Abhinandan Krishnan, Luke Zettlemoyer, and Madian Khabsa. 2023.
\newblock \href {https://arxiv.org/abs/2308.16884} {The belebele benchmark: a parallel reading comprehension dataset in 122 language variants}.
\newblock \emph{Preprint}, arXiv:2308.16884.

\bibitem[{Biderman et~al.(2024)Biderman, Ortiz, Portes, Paul, Greengard, Jennings, King, Havens, Chiley, Frankle et~al.}]{biderman2024lora}
Dan Biderman, Jose~Gonzalez Ortiz, Jacob Portes, Mansheej Paul, Philip Greengard, Connor Jennings, Daniel King, Sam Havens, Vitaliy Chiley, Jonathan Frankle, et~al. 2024.
\newblock Lora learns less and forgets less.
\newblock \emph{arXiv preprint arXiv:2405.09673}.

\bibitem[{Chaudhary(2023)}]{codealpaca}
S.~Chaudhary. 2023.
\newblock \href {https://github.com/sahil280114/codealpaca.} {Code alpaca: An instruction-following llama model for code generation}.

\bibitem[{Chen et~al.(2023)Chen, Yan, Liang, Jiang, Wu, Yu, Chen, Chen, Zhang, Jianquan, Xiang, and Wang}]{Chen_MultilingualSIFT_Multilingual_Supervised_2023}
Zhihong Chen, Shuo Yan, Juhao Liang, Feng Jiang, Xiangbo Wu, Fei Yu, Guiming~Hardy Chen, Junying Chen, Hongbo Zhang, Li~Jianquan, Wan Xiang, and Benyou Wang. 2023.
\newblock \href {https://github.com/FreedomIntelligence/MultilingualSIFT.git} {{MultilingualSIFT: Multilingual Supervised Instruction Fine-tuning}}.

\bibitem[{Choshen et~al.(2022)Choshen, Venezian, Slonim, and Katz}]{choshen2022fusing}
Leshem Choshen, Elad Venezian, Noam Slonim, and Yoav Katz. 2022.
\newblock \href {https://arxiv.org/abs/2204.03044} {Fusing finetuned models for better pretraining}.
\newblock \emph{Preprint}, arXiv:2204.03044.

\bibitem[{Clark et~al.(2018)Clark, Cowhey, Etzioni, Khot, Sabharwal, Schoenick, and Tafjord}]{arc}
Peter Clark, Isaac Cowhey, Oren Etzioni, Tushar Khot, Ashish Sabharwal, Carissa Schoenick, and Oyvind Tafjord. 2018.
\newblock \href {https://api.semanticscholar.org/CorpusID:3922816} {Think you have solved question answering? try arc, the ai2 reasoning challenge}.
\newblock \emph{ArXiv}, abs/1803.05457.

\bibitem[{Cobbe et~al.(2021)Cobbe, Kosaraju, Bavarian, Chen, Jun, Kaiser, Plappert, Tworek, Hilton, Nakano, Hesse, and Schulman}]{gsm8k}
Karl Cobbe, Vineet Kosaraju, Mohammad Bavarian, Mark Chen, Heewoo Jun, Lukasz Kaiser, Matthias Plappert, Jerry Tworek, Jacob Hilton, Reiichiro Nakano, Christopher Hesse, and John Schulman. 2021.
\newblock \href {https://api.semanticscholar.org/CorpusID:239998651} {Training verifiers to solve math word problems}.
\newblock \emph{ArXiv}, abs/2110.14168.

\bibitem[{Conneau et~al.(2018)Conneau, Rinott, Lample, Williams, Bowman, Schwenk, and Stoyanov}]{xnli}
Alexis Conneau, Ruty Rinott, Guillaume Lample, Adina Williams, Samuel Bowman, Holger Schwenk, and Veselin Stoyanov. 2018.
\newblock \href {https://doi.org/10.18653/v1/D18-1269} {{XNLI}: Evaluating cross-lingual sentence representations}.
\newblock In \emph{Proceedings of the 2018 Conference on Empirical Methods in Natural Language Processing}, pages 2475--2485, Brussels, Belgium. Association for Computational Linguistics.

\bibitem[{Cui and Yao(2024)}]{CuiY24}
Yiming Cui and Xin Yao. 2024.
\newblock \href {https://doi.org/10.48550/ARXIV.2403.01851} {Rethinking {LLM} language adaptation: {A} case study on chinese mixtral}.
\newblock \emph{CoRR}, abs/2403.01851.

\bibitem[{Daniele and Suphavadeeprasit(2023)}]{capybara}
Luigi Daniele and Suphavadeeprasit. 2023.
\newblock \href {https://huggingface.co/datasets/LDJnr/Capybara} {Amplify-instruct: Synthetically generated diverse multi-turn conversations for efficient llm training.}
\newblock \emph{arXiv preprint arXiv:(coming soon)}.

\bibitem[{Dao(2024)}]{dao2024flashattention}
Tri Dao. 2024.
\newblock \href {https://openreview.net/forum?id=mZn2Xyh9Ec} {Flashattention-2: Faster attention with better parallelism and work partitioning}.
\newblock In \emph{The Twelfth International Conference on Learning Representations}.

\bibitem[{Davari and Belilovsky(2023)}]{DavariB23}
MohammadReza Davari and Eugene Belilovsky. 2023.
\newblock \href {https://doi.org/10.48550/ARXIV.2312.06795} {Model breadcrumbs: Scaling multi-task model merging with sparse masks}.
\newblock \emph{CoRR}, abs/2312.06795.

\bibitem[{Don-Yehiya et~al.(2023)Don-Yehiya, Venezian, Raffel, Slonim, and Choshen}]{cold}
Shachar Don-Yehiya, Elad Venezian, Colin Raffel, Noam Slonim, and Leshem Choshen. 2023.
\newblock \href {https://doi.org/10.18653/v1/2023.acl-long.46} {{C}ol{D} fusion: Collaborative descent for distributed multitask finetuning}.
\newblock In \emph{Proceedings of the 61st Annual Meeting of the Association for Computational Linguistics (Volume 1: Long Papers)}, pages 788--806, Toronto, Canada. Association for Computational Linguistics.

\bibitem[{Erjavec et~al.(2023)Erjavec, Ogrodniczuk, Osenova et~al.}]{parlamint}
Toma{\v{z}} Erjavec, Maciej Ogrodniczuk, Petya Osenova, et~al. 2023.
\newblock \href {https://doi.org/10.1007/s10579-021-09574-0} {The parlamint corpora of parliamentary proceedings}.
\newblock \emph{Language Resources and Evaluation}, 57(2):415--448.

\bibitem[{Foundation()}]{wikidump}
Wikimedia Foundation.
\newblock \href {https://dumps.wikimedia.org} {Wikimedia downloads}.

\bibitem[{French(1999)}]{french1999catastrophic}
Robert~M French. 1999.
\newblock Catastrophic forgetting in connectionist networks.
\newblock \emph{Trends in cognitive sciences}, 3(4):128--135.

\bibitem[{Fu and Khot(2022)}]{fu2022gptroadmap}
Hao Fu, Yao;~Peng and Tushar Khot. 2022.
\newblock \href {https://yaofu.notion.site/How-does-GPT-Obtain-its-Ability-Tracing-Emergent-Abilities-of-Language-Models-to-their-Sources-b9a57ac0fcf74f30a1ab9e3e36fa1dc1} {How does gpt obtain its ability? tracing emergent abilities of language models to their sources}.
\newblock \emph{Yao Fu’s Notion}.

\bibitem[{Gao et~al.(2023)Gao, Tow, Abbasi, Biderman, Black, DiPofi, Foster, Golding, Hsu, Le~Noac'h, Li, McDonell, Muennighoff, Ociepa, Phang, Reynolds, Schoelkopf, Skowron, Sutawika, Tang, Thite, Wang, Wang, and Zou}]{eval-harness}
Leo Gao, Jonathan Tow, Baber Abbasi, Stella Biderman, Sid Black, Anthony DiPofi, Charles Foster, Laurence Golding, Jeffrey Hsu, Alain Le~Noac'h, Haonan Li, Kyle McDonell, Niklas Muennighoff, Chris Ociepa, Jason Phang, Laria Reynolds, Hailey Schoelkopf, Aviya Skowron, Lintang Sutawika, Eric Tang, Anish Thite, Ben Wang, Kevin Wang, and Andy Zou. 2023.
\newblock \href {https://doi.org/10.5281/zenodo.10256836} {A framework for few-shot language model evaluation}.

\bibitem[{Gee et~al.(2022)Gee, Zugarini, Rigutini, and Torroni}]{fvt}
Leonidas Gee, Andrea Zugarini, Leonardo Rigutini, and Paolo Torroni. 2022.
\newblock \href {https://doi.org/10.18653/v1/2022.emnlp-industry.41} {Fast vocabulary transfer for language model compression}.
\newblock In \emph{Proceedings of the 2022 Conference on Empirical Methods in Natural Language Processing: Industry Track}, pages 409--416, Abu Dhabi, UAE. Association for Computational Linguistics.

\bibitem[{Goddard et~al.(2024)Goddard, Siriwardhana, Ehghaghi, Meyers, Karpukhin, Benedict, McQuade, and Solawetz}]{goddard2024arcee}
Charles Goddard, Shamane Siriwardhana, Malikeh Ehghaghi, Luke Meyers, Vlad Karpukhin, Brian Benedict, Mark McQuade, and Jacob Solawetz. 2024.
\newblock Arcee's mergekit: A toolkit for merging large language models.
\newblock \emph{arXiv preprint arXiv:2403.13257}.

\bibitem[{Gogoulou et~al.(2023)Gogoulou, Lesort, Boman, and Nivre}]{GogoulouLBN23}
Evangelia Gogoulou, Timoth{\'{e}}e Lesort, Magnus Boman, and Joakim Nivre. 2023.
\newblock \href {https://doi.org/10.48550/ARXIV.2311.01200} {A study of continual learning under language shift}.
\newblock \emph{CoRR}, abs/2311.01200.

\bibitem[{Gokaslan et~al.(2019)Gokaslan, Cohen, Pavlick, and Tellex}]{openwebtext}
Aaron Gokaslan, Vanya Cohen, Ellie Pavlick, and Stefanie Tellex. 2019.
\newblock Openwebtext corpus.
\newblock \url{http://Skylion007.github.io/OpenWebTextCorpus}.

\bibitem[{Goodfellow et~al.(2014)Goodfellow, Mirza, Da, Courville, and Bengio}]{GoodfellowMDCB13}
Ian~J. Goodfellow, Mehdi Mirza, Xia Da, Aaron~C. Courville, and Yoshua Bengio. 2014.
\newblock \href {http://arxiv.org/abs/1312.6211} {An empirical investigation of catastrophic forgeting in gradient-based neural networks}.
\newblock In \emph{2nd International Conference on Learning Representations, {ICLR} 2014, Banff, AB, Canada, April 14-16, 2014, Conference Track Proceedings}.

\bibitem[{Groeneveld et~al.(2024)Groeneveld, Beltagy, Walsh, Bhagia, Kinney, Tafjord, Jha, Ivison, Magnusson, Wang, Arora, Atkinson, Authur, Chandu, Cohan, Dumas, Elazar, Gu, Hessel, Khot, Merrill, Morrison, Muennighoff, Naik, Nam, Peters, Pyatkin, Ravichander, Schwenk, Shah, Smith, Strubell, Subramani, Wortsman, Dasigi, Lambert, Richardson, Zettlemoyer, Dodge, Lo, Soldaini, Smith, and Hajishirzi}]{olmo}
Dirk Groeneveld, Iz~Beltagy, Pete Walsh, Akshita Bhagia, Rodney Kinney, Oyvind Tafjord, Ananya~Harsh Jha, Hamish Ivison, Ian Magnusson, Yizhong Wang, Shane Arora, David Atkinson, Russell Authur, Khyathi~Raghavi Chandu, Arman Cohan, Jennifer Dumas, Yanai Elazar, Yuling Gu, Jack Hessel, Tushar Khot, William Merrill, Jacob Morrison, Niklas Muennighoff, Aakanksha Naik, Crystal Nam, Matthew~E. Peters, Valentina Pyatkin, Abhilasha Ravichander, Dustin Schwenk, Saurabh Shah, Will Smith, Emma Strubell, Nishant Subramani, Mitchell Wortsman, Pradeep Dasigi, Nathan Lambert, Kyle Richardson, Luke Zettlemoyer, Jesse Dodge, Kyle Lo, Luca Soldaini, Noah~A. Smith, and Hannaneh Hajishirzi. 2024.
\newblock \href {https://doi.org/10.48550/ARXIV.2402.00838} {Olmo: Accelerating the science of language models}.
\newblock \emph{CoRR}, abs/2402.00838.

\bibitem[{Hardalov et~al.(2020)Hardalov, Mihaylov, Zlatkova, Dinkov, Koychev, and Nakov}]{exams}
Momchil Hardalov, Todor Mihaylov, Dimitrina Zlatkova, Yoan Dinkov, Ivan Koychev, and Preslav Nakov. 2020.
\newblock \href {https://doi.org/10.18653/v1/2020.emnlp-main.438} {{EXAMS}: A multi-subject high school examinations dataset for cross-lingual and multilingual question answering}.
\newblock In \emph{Proceedings of the 2020 Conference on Empirical Methods in Natural Language Processing (EMNLP)}, pages 5427--5444, Online. Association for Computational Linguistics.

\bibitem[{Hendrycks et~al.(2021)Hendrycks, Burns, Basart, Zou, Mazeika, Song, and Steinhardt}]{mmlu}
Dan Hendrycks, Collin Burns, Steven Basart, Andy Zou, Mantas Mazeika, Dawn Song, and Jacob Steinhardt. 2021.
\newblock \href {https://openreview.net/forum?id=d7KBjmI3GmQ} {Measuring massive multitask language understanding}.
\newblock In \emph{International Conference on Learning Representations}.

\bibitem[{Hu et~al.(2022)Hu, yelong shen, Wallis, Allen-Zhu, Li, Wang, Wang, and Chen}]{hu2022lora}
Edward~J Hu, yelong shen, Phillip Wallis, Zeyuan Allen-Zhu, Yuanzhi Li, Shean Wang, Lu~Wang, and Weizhu Chen. 2022.
\newblock \href {https://openreview.net/forum?id=nZeVKeeFYf9} {Lo{RA}: Low-rank adaptation of large language models}.
\newblock In \emph{International Conference on Learning Representations}.

\bibitem[{Huang et~al.(2024)Huang, Cui, Wang, Yang, Liao, Song, Yao, and Su}]{HuangCWYLSYS24}
Jianheng Huang, Leyang Cui, Ante Wang, Chengyi Yang, Xinting Liao, Linfeng Song, Junfeng Yao, and Jinsong Su. 2024.
\newblock \href {https://doi.org/10.48550/ARXIV.2403.01244} {Mitigating catastrophic forgetting in large language models with self-synthesized rehearsal}.
\newblock \emph{CoRR}, abs/2403.01244.

\bibitem[{Ibrahim et~al.(2024)Ibrahim, Thérien, Gupta, Richter, Anthony, Lesort, Belilovsky, and Rish}]{simple}
Adam Ibrahim, Benjamin Thérien, Kshitij Gupta, Mats~L. Richter, Quentin Anthony, Timothée Lesort, Eugene Belilovsky, and Irina Rish. 2024.
\newblock \href {https://openreview.net/forum?id=DimPeeCxKO} {Simple and scalable strategies to continually pre-train large language models}.
\newblock \emph{Submitted to Transactions on Machine Learning Research}.
\newblock Under review.

\bibitem[{Ilharco et~al.(2023)Ilharco, Ribeiro, Wortsman, Schmidt, Hajishirzi, and Farhadi}]{IlharcoRWSHF23}
Gabriel Ilharco, Marco~T{\'{u}}lio Ribeiro, Mitchell Wortsman, Ludwig Schmidt, Hannaneh Hajishirzi, and Ali Farhadi. 2023.
\newblock \href {https://openreview.net/pdf?id=6t0Kwf8-jrj} {Editing models with task arithmetic}.
\newblock In \emph{The Eleventh International Conference on Learning Representations, {ICLR} 2023, Kigali, Rwanda, May 1-5, 2023}. OpenReview.net.

\bibitem[{Jain et~al.(2024)Jain, yeh Chiang, Wen, Kirchenbauer, Chu, Somepalli, Bartoldson, Kailkhura, Schwarzschild, Saha, Goldblum, Geiping, and Goldstein}]{jain2024neftune}
Neel Jain, Ping yeh Chiang, Yuxin Wen, John Kirchenbauer, Hong-Min Chu, Gowthami Somepalli, Brian~R. Bartoldson, Bhavya Kailkhura, Avi Schwarzschild, Aniruddha Saha, Micah Goldblum, Jonas Geiping, and Tom Goldstein. 2024.
\newblock \href {https://openreview.net/forum?id=0bMmZ3fkCk} {{NEFT}une: Noisy embeddings improve instruction finetuning}.
\newblock In \emph{The Twelfth International Conference on Learning Representations}.

\bibitem[{Jang et~al.(2024)Jang, Yun, and Han}]{JangYH24}
Dong{-}Hwan Jang, Sangdoo Yun, and Dongyoon Han. 2024.
\newblock \href {https://doi.org/10.48550/ARXIV.2403.19522} {Model stock: All we need is just a few fine-tuned models}.
\newblock \emph{CoRR}, abs/2403.19522.

\bibitem[{Jiang et~al.(2023)Jiang, Sablayrolles, Mensch, Bamford, Chaplot, de~Las~Casas, Bressand, Lengyel, Lample, Saulnier, Lavaud, Lachaux, Stock, Scao, Lavril, Wang, Lacroix, and Sayed}]{Mistral7B}
Albert~Q. Jiang, Alexandre Sablayrolles, Arthur Mensch, Chris Bamford, Devendra~Singh Chaplot, Diego de~Las~Casas, Florian Bressand, Gianna Lengyel, Guillaume Lample, Lucile Saulnier, L{\'{e}}lio~Renard Lavaud, Marie{-}Anne Lachaux, Pierre Stock, Teven~Le Scao, Thibaut Lavril, Thomas Wang, Timoth{\'{e}}e Lacroix, and William~El Sayed. 2023.
\newblock \href {https://doi.org/10.48550/ARXIV.2310.06825} {Mistral 7b}.
\newblock \emph{CoRR}, abs/2310.06825.

\bibitem[{Jiang et~al.(2024)Jiang, Sun, Shi, Rodriguez, Zhou, Neubig, Lin, Yih, and Iyer}]{JiangSSRZNLYI24}
Zhengbao Jiang, Zhiqing Sun, Weijia Shi, Pedro Rodriguez, Chunting Zhou, Graham Neubig, Xi~Victoria Lin, Wen{-}tau Yih, and Srinivasan Iyer. 2024.
\newblock \href {https://doi.org/10.48550/ARXIV.2402.12847} {Instruction-tuned language models are better knowledge learners}.
\newblock \emph{CoRR}, abs/2402.12847.

\bibitem[{Joshi et~al.(2017)Joshi, Choi, Weld, and Zettlemoyer}]{triviaqa}
Mandar Joshi, Eunsol Choi, Daniel Weld, and Luke Zettlemoyer. 2017.
\newblock \href {https://doi.org/10.18653/v1/P17-1147} {{T}rivia{QA}: A large scale distantly supervised challenge dataset for reading comprehension}.
\newblock In \emph{Proceedings of the 55th Annual Meeting of the Association for Computational Linguistics (Volume 1: Long Papers)}, pages 1601--1611, Vancouver, Canada. Association for Computational Linguistics.

\bibitem[{Kemker et~al.(2018)Kemker, McClure, Abitino, Hayes, and Kanan}]{KemkerMAHK18}
Ronald Kemker, Marc McClure, Angelina Abitino, Tyler~L. Hayes, and Christopher Kanan. 2018.
\newblock \href {https://doi.org/10.1609/AAAI.V32I1.11651} {Measuring catastrophic forgetting in neural networks}.
\newblock In \emph{Proceedings of the Thirty-Second {AAAI} Conference on Artificial Intelligence, (AAAI-18), the 30th innovative Applications of Artificial Intelligence (IAAI-18), and the 8th {AAAI} Symposium on Educational Advances in Artificial Intelligence (EAAI-18), New Orleans, Louisiana, USA, February 2-7, 2018}, pages 3390--3398. {AAAI} Press.

\bibitem[{Koehn(2005)}]{koehn-2005-europarl}
Philipp Koehn. 2005.
\newblock \href {https://aclanthology.org/2005.mtsummit-papers.11} {{E}uroparl: A parallel corpus for statistical machine translation}.
\newblock In \emph{Proceedings of Machine Translation Summit X: Papers}, pages 79--86, Phuket, Thailand.

\bibitem[{Kudo and Richardson(2018)}]{sentencepiece}
Taku Kudo and John Richardson. 2018.
\newblock \href {https://doi.org/10.18653/v1/D18-2012} {{S}entence{P}iece: A simple and language independent subword tokenizer and detokenizer for neural text processing}.
\newblock In \emph{Proceedings of the 2018 Conference on Empirical Methods in Natural Language Processing: System Demonstrations}, pages 66--71, Brussels, Belgium. Association for Computational Linguistics.

\bibitem[{Lee et~al.(2023)Lee, Hunter, and Ruiz}]{platypus2023}
Ariel~N. Lee, Cole~J. Hunter, and Nataniel Ruiz. 2023.
\newblock Platypus: Quick, cheap, and powerful refinement of llms.

\bibitem[{Lee et~al.(2020)Lee, Cho, and Kang}]{LeeCK20}
Cheolhyoung Lee, Kyunghyun Cho, and Wanmo Kang. 2020.
\newblock \href {https://openreview.net/forum?id=HkgaETNtDB} {Mixout: Effective regularization to finetune large-scale pretrained language models}.
\newblock In \emph{8th International Conference on Learning Representations, {ICLR} 2020, Addis Ababa, Ethiopia, April 26-30, 2020}. OpenReview.net.

\bibitem[{Li and Lee(2024)}]{LiL24}
Chen{-}An Li and Hung{-}Yi Lee. 2024.
\newblock \href {https://doi.org/10.48550/ARXIV.2401.03129} {Examining forgetting in continual pre-training of aligned large language models}.
\newblock \emph{CoRR}, abs/2401.03129.

\bibitem[{Li et~al.(2022)Li, Gururangan, Dettmers, Lewis, Althoff, Smith, and Zettlemoyer}]{btm}
Margaret Li, Suchin Gururangan, Tim Dettmers, Mike Lewis, Tim Althoff, Noah~A. Smith, and Luke Zettlemoyer. 2022.
\newblock \href {https://openreview.net/forum?id=SQgVgE2Sq4} {Branch-train-merge: Embarrassingly parallel training of expert language models}.
\newblock In \emph{First Workshop on Interpolation Regularizers and Beyond at NeurIPS 2022}.

\bibitem[{Lian et~al.(2023)Lian, Wang, Goodson, Pentland, Cook, Vong, and "Teknium"}]{slimorca}
Wing Lian, Guan Wang, Bleys Goodson, Eugene Pentland, Austin Cook, Chanvichet Vong, and "Teknium". 2023.
\newblock \href {https://https://huggingface.co/Open-Orca/SlimOrca} {Slimorca: An open dataset of gpt-4 augmented flan reasoning traces, with verification}.

\bibitem[{Liang et~al.(2022)Liang, Bommasani, Lee, Tsipras, Soylu, Yasunaga, Zhang, Narayanan, Wu, Kumar, Newman, Yuan, Yan, Zhang, Cosgrove, Manning, R{\'{e}}, Acosta{-}Navas, Hudson, Zelikman, Durmus, Ladhak, Rong, Ren, Yao, Wang, Santhanam, Orr, Zheng, Y{\"{u}}ksekg{\"{o}}n{\"{u}}l, Suzgun, Kim, Guha, Chatterji, Khattab, Henderson, Huang, Chi, Xie, Santurkar, Ganguli, Hashimoto, Icard, Zhang, Chaudhary, Wang, Li, Mai, Zhang, and Koreeda}]{Liang23}
Percy Liang, Rishi Bommasani, Tony Lee, Dimitris Tsipras, Dilara Soylu, Michihiro Yasunaga, Yian Zhang, Deepak Narayanan, Yuhuai Wu, Ananya Kumar, Benjamin Newman, Binhang Yuan, Bobby Yan, Ce~Zhang, Christian Cosgrove, Christopher~D. Manning, Christopher R{\'{e}}, Diana Acosta{-}Navas, Drew~A. Hudson, Eric Zelikman, Esin Durmus, Faisal Ladhak, Frieda Rong, Hongyu Ren, Huaxiu Yao, Jue Wang, Keshav Santhanam, Laurel~J. Orr, Lucia Zheng, Mert Y{\"{u}}ksekg{\"{o}}n{\"{u}}l, Mirac Suzgun, Nathan Kim, Neel Guha, Niladri~S. Chatterji, Omar Khattab, Peter Henderson, Qian Huang, Ryan Chi, Sang~Michael Xie, Shibani Santurkar, Surya Ganguli, Tatsunori Hashimoto, Thomas Icard, Tianyi Zhang, Vishrav Chaudhary, William Wang, Xuechen Li, Yifan Mai, Yuhui Zhang, and Yuta Koreeda. 2022.
\newblock \href {https://doi.org/10.48550/ARXIV.2211.09110} {Holistic evaluation of language models}.
\newblock \emph{CoRR}, abs/2211.09110.

\bibitem[{Lin et~al.(2024)Lin, Lin, Xiong, Diao, Liu, Zhang, Pan, Wang, Hu, Zhang, Dong, Pi, Zhao, Jiang, Ji, Yao, and Zhang}]{lin2024mitigating}
Yong Lin, Hangyu Lin, Wei Xiong, Shizhe Diao, Jianmeng Liu, Jipeng Zhang, Rui Pan, Haoxiang Wang, Wenbin Hu, Hanning Zhang, Hanze Dong, Renjie Pi, Han Zhao, Nan Jiang, Heng Ji, Yuan Yao, and Tong Zhang. 2024.
\newblock \href {https://arxiv.org/abs/2309.06256} {Mitigating the alignment tax of rlhf}.
\newblock \emph{Preprint}, arXiv:2309.06256.

\bibitem[{Longpre et~al.(2023)Longpre, Hou, Vu, Webson, Chung, Tay, Zhou, Le, Zoph, Wei, and Roberts}]{theflancollection}
Shayne Longpre, Le~Hou, Tu~Vu, Albert Webson, Hyung~Won Chung, Yi~Tay, Denny Zhou, Quoc~V Le, Barret Zoph, Jason Wei, and Adam Roberts. 2023.
\newblock \href {https://proceedings.mlr.press/v202/longpre23a.html} {The flan collection: Designing data and methods for effective instruction tuning}.
\newblock In \emph{Proceedings of the 40th International Conference on Machine Learning}, volume 202 of \emph{Proceedings of Machine Learning Research}, pages 22631--22648. PMLR.

\bibitem[{Lozhkov et~al.(2024)Lozhkov, Ben~Allal, von Werra, and Wolf}]{fineweb-edu}
Anton Lozhkov, Loubna Ben~Allal, Leandro von Werra, and Thomas Wolf. 2024.
\newblock \href {https://doi.org/10.57967/hf/2497} {Fineweb-edu}.

\bibitem[{Ma et~al.(2023)Ma, Liu, Yu, Zhang, Jiang, Wang, and Li}]{MaLYZJWL23}
Yingwei Ma, Yue Liu, Yue Yu, Yuanliang Zhang, Yu~Jiang, Changjian Wang, and Shanshan Li. 2023.
\newblock \href {https://doi.org/10.48550/ARXIV.2309.16298} {At which training stage does code data help llms reasoning?}
\newblock \emph{CoRR}, abs/2309.16298.

\bibitem[{Matena and Raffel(2022)}]{MatenaR22}
Michael Matena and Colin Raffel. 2022.
\newblock \href {http://papers.nips.cc/paper\_files/paper/2022/hash/70c26937fbf3d4600b69a129031b66ec-Abstract-Conference.html} {Merging models with fisher-weighted averaging}.
\newblock In \emph{Advances in Neural Information Processing Systems 35: Annual Conference on Neural Information Processing Systems 2022, NeurIPS 2022, New Orleans, LA, USA, November 28 - December 9, 2022}.

\bibitem[{McMahan et~al.(2017)McMahan, Moore, Ramage, Hampson, and y~Arcas}]{McMahanMRHA17}
Brendan McMahan, Eider Moore, Daniel Ramage, Seth Hampson, and Blaise~Ag{\"{u}}era y~Arcas. 2017.
\newblock \href {http://proceedings.mlr.press/v54/mcmahan17a.html} {Communication-efficient learning of deep networks from decentralized data}.
\newblock In \emph{Proceedings of the 20th International Conference on Artificial Intelligence and Statistics, {AISTATS} 2017, 20-22 April 2017, Fort Lauderdale, FL, {USA}}, volume~54 of \emph{Proceedings of Machine Learning Research}, pages 1273--1282. {PMLR}.

\bibitem[{Mitra et~al.(2024)Mitra, Khanpour, Rosset, and Awadallah}]{mitra2024orcamath}
Arindam Mitra, Hamed Khanpour, Corby Rosset, and Ahmed Awadallah. 2024.
\newblock \href {https://arxiv.org/abs/2402.14830} {Orca-math: Unlocking the potential of slms in grade school math}.
\newblock \emph{Preprint}, arXiv:2402.14830.

\bibitem[{Mosin et~al.(2023)Mosin, Samenko, Kozlovskii, Tikhonov, and Yamshchikov}]{vipi}
Vladislav Mosin, Igor Samenko, Borislav Kozlovskii, Alexey Tikhonov, and Ivan~P. Yamshchikov. 2023.
\newblock \href {https://doi.org/10.1016/j.artint.2023.103860} {Fine-tuning transformers: Vocabulary transfer}.
\newblock \emph{Artif. Intell.}, 317(C).

\bibitem[{Mukherjee et~al.(2023)Mukherjee, Mitra, Jawahar, Agarwal, Palangi, and Awadallah}]{mukherjee2023orca}
Subhabrata Mukherjee, Arindam Mitra, Ganesh Jawahar, Sahaj Agarwal, Hamid Palangi, and Ahmed Awadallah. 2023.
\newblock \href {https://arxiv.org/abs/2306.02707} {Orca: Progressive learning from complex explanation traces of gpt-4}.
\newblock \emph{Preprint}, arXiv:2306.02707.

\bibitem[{Namata et~al.(2012)Namata, London, Getoor, and Huang}]{pubmed}
Galileo~Mark Namata, Ben London, Lise Getoor, and Bert Huang. 2012.
\newblock Query-driven active surveying for collective classification.
\newblock In \emph{International Workshop on Mining and Learning with Graphs}, Edinburgh, Scotland.

\bibitem[{Nguyen et~al.(2024)Nguyen, Nguyen, Lai, Man, Ngo, Dernoncourt, Rossi, and Nguyen}]{culturax}
Thuat Nguyen, Chien~Van Nguyen, Viet~Dac Lai, Hieu Man, Nghia~Trung Ngo, Franck Dernoncourt, Ryan~A. Rossi, and Thien~Huu Nguyen. 2024.
\newblock \href {https://aclanthology.org/2024.lrec-main.377} {{C}ultura{X}: A cleaned, enormous, and multilingual dataset for large language models in 167 languages}.
\newblock In \emph{Proceedings of the 2024 Joint International Conference on Computational Linguistics, Language Resources and Evaluation (LREC-COLING 2024)}, pages 4226--4237, Torino, Italia. ELRA and ICCL.

\bibitem[{Paszke et~al.(2019)Paszke, Gross, Massa, Lerer, Bradbury, Chanan, Killeen, Lin, Gimelshein, Antiga, Desmaison, K{\"{o}}pf, Yang, DeVito, Raison, Tejani, Chilamkurthy, Steiner, Fang, Bai, and Chintala}]{PaszkeGMLBCKLGA19}
Adam Paszke, Sam Gross, Francisco Massa, Adam Lerer, James Bradbury, Gregory Chanan, Trevor Killeen, Zeming Lin, Natalia Gimelshein, Luca Antiga, Alban Desmaison, Andreas K{\"{o}}pf, Edward~Z. Yang, Zachary DeVito, Martin Raison, Alykhan Tejani, Sasank Chilamkurthy, Benoit Steiner, Lu~Fang, Junjie Bai, and Soumith Chintala. 2019.
\newblock \href {https://proceedings.neurips.cc/paper/2019/hash/bdbca288fee7f92f2bfa9f7012727740-Abstract.html} {Pytorch: An imperative style, high-performance deep learning library}.
\newblock In \emph{Advances in Neural Information Processing Systems 32: Annual Conference on Neural Information Processing Systems 2019, NeurIPS 2019, December 8-14, 2019, Vancouver, BC, Canada}, pages 8024--8035.

\bibitem[{Penedo et~al.(2023)Penedo, Malartic, Hesslow, Cojocaru, Alobeidli, Cappelli, Pannier, Almazrouei, and Launay}]{PenedoMHCACPAL23}
Guilherme Penedo, Quentin Malartic, Daniel Hesslow, Ruxandra Cojocaru, Hamza Alobeidli, Alessandro Cappelli, Baptiste Pannier, Ebtesam Almazrouei, and Julien Launay. 2023.
\newblock \href {http://papers.nips.cc/paper\_files/paper/2023/hash/fa3ed726cc5073b9c31e3e49a807789c-Abstract-Datasets\_and\_Benchmarks.html} {The refinedweb dataset for falcon {LLM:} outperforming curated corpora with web data only}.
\newblock In \emph{Advances in Neural Information Processing Systems 36: Annual Conference on Neural Information Processing Systems 2023, NeurIPS 2023, New Orleans, LA, USA, December 10 - 16, 2023}.

\bibitem[{Plüster(2023{\natexlab{a}})}]{german_benchmarks}
Björn Plüster. 2023{\natexlab{a}}.
\newblock \href {https://github.com/bjoernpl/GermanBenchmark} {{German Benchmark Datasets}}.

\bibitem[{Plüster(2023{\natexlab{b}})}]{openschnabeltier}
Björn Plüster. 2023{\natexlab{b}}.
\newblock \href {https://huggingface.co/datasets/LeoLM/OpenSchnabeltier} {Leolm/openschnabeltier dataset}.

\bibitem[{Radford et~al.(2019)Radford, Wu, Child, Luan, Amodei, and Sutskever}]{gpt2}
Alec Radford, Jeff Wu, Rewon Child, David Luan, Dario Amodei, and Ilya Sutskever. 2019.
\newblock \href {https://api.semanticscholar.org/CorpusID:160025533} {Language models are unsupervised multitask learners}.

\bibitem[{Rajbhandari et~al.(2020{\natexlab{a}})Rajbhandari, Rasley, Ruwase, and He}]{zero9355301}
Samyam Rajbhandari, Jeff Rasley, Olatunji Ruwase, and Yuxiong He. 2020{\natexlab{a}}.
\newblock \href {https://doi.org/10.1109/SC41405.2020.00024} {Zero: Memory optimizations toward training trillion parameter models}.
\newblock In \emph{SC20: International Conference for High Performance Computing, Networking, Storage and Analysis}, pages 1--16.

\bibitem[{Rajbhandari et~al.(2020{\natexlab{b}})Rajbhandari, Rasley, Ruwase, and He}]{dszero}
Samyam Rajbhandari, Jeff Rasley, Olatunji Ruwase, and Yuxiong He. 2020{\natexlab{b}}.
\newblock Zero: memory optimizations toward training trillion parameter models.
\newblock In \emph{Proceedings of the International Conference for High Performance Computing, Networking, Storage and Analysis}, SC '20. IEEE Press.

\bibitem[{Ramasesh et~al.(2022)Ramasesh, Lewkowycz, and Dyer}]{RamaseshLD22}
Vinay~Venkatesh Ramasesh, Aitor Lewkowycz, and Ethan Dyer. 2022.
\newblock \href {https://openreview.net/forum?id=GhVS8\_yPeEa} {Effect of scale on catastrophic forgetting in neural networks}.
\newblock In \emph{The Tenth International Conference on Learning Representations, {ICLR} 2022, Virtual Event, April 25-29, 2022}. OpenReview.net.

\bibitem[{Rasley et~al.(2020)Rasley, Rajbhandari, Ruwase, and He}]{deepspeed10.1145/3394486.3406703}
Jeff Rasley, Samyam Rajbhandari, Olatunji Ruwase, and Yuxiong He. 2020.
\newblock \href {https://doi.org/10.1145/3394486.3406703} {Deepspeed: System optimizations enable training deep learning models with over 100 billion parameters}.
\newblock In \emph{Proceedings of the 26th ACM SIGKDD International Conference on Knowledge Discovery \& Data Mining}, KDD '20, page 3505–3506, New York, NY, USA. Association for Computing Machinery.

\bibitem[{Rolnick et~al.(2019)Rolnick, Ahuja, Schwarz, Lillicrap, and Wayne}]{RolnickASLW19}
David Rolnick, Arun Ahuja, Jonathan Schwarz, Timothy~P. Lillicrap, and Gregory Wayne. 2019.
\newblock \href {https://proceedings.neurips.cc/paper/2019/hash/fa7cdfad1a5aaf8370ebeda47a1ff1c3-Abstract.html} {Experience replay for continual learning}.
\newblock In \emph{Advances in Neural Information Processing Systems 32: Annual Conference on Neural Information Processing Systems 2019, NeurIPS 2019, December 8-14, 2019, Vancouver, BC, Canada}, pages 348--358.

\bibitem[{Sakaguchi et~al.(2021)Sakaguchi, Bras, Bhagavatula, and Choi}]{winogrande}
Keisuke Sakaguchi, Ronan~Le Bras, Chandra Bhagavatula, and Yejin Choi. 2021.
\newblock \href {https://doi.org/10.1145/3474381} {Winogrande: an adversarial winograd schema challenge at scale}.
\newblock \emph{Commun. ACM}, 64(9):99–106.

\bibitem[{Scialom et~al.(2022)Scialom, Chakrabarty, and Muresan}]{ScialomCM22}
Thomas Scialom, Tuhin Chakrabarty, and Smaranda Muresan. 2022.
\newblock \href {https://doi.org/10.18653/V1/2022.EMNLP-MAIN.410} {Fine-tuned language models are continual learners}.
\newblock In \emph{Proceedings of the 2022 Conference on Empirical Methods in Natural Language Processing, {EMNLP} 2022, Abu Dhabi, United Arab Emirates, December 7-11, 2022}, pages 6107--6122. Association for Computational Linguistics.

\bibitem[{Shen et~al.(2023)Shen, Tao, Ma, Neiswanger, Liu, Wang, Tan, Hestness, Vassilieva, Soboleva, and Xing}]{ShenTMNLWTHVSX23}
Zhiqiang Shen, Tianhua Tao, Liqun Ma, Willie Neiswanger, Zhengzhong Liu, Hongyi Wang, Bowen Tan, Joel Hestness, Natalia Vassilieva, Daria Soboleva, and Eric~P. Xing. 2023.
\newblock \href {https://doi.org/10.48550/ARXIV.2309.10818} {Slimpajama-dc: Understanding data combinations for {LLM} training}.
\newblock \emph{CoRR}, abs/2309.10818.

\bibitem[{Shi et~al.(2023)Shi, Suzgun, Freitag, Wang, Srivats, Vosoughi, Chung, Tay, Ruder, Zhou, Das, and Wei}]{mgsm}
Freda Shi, Mirac Suzgun, Markus Freitag, Xuezhi Wang, Suraj Srivats, Soroush Vosoughi, Hyung~Won Chung, Yi~Tay, Sebastian Ruder, Denny Zhou, Dipanjan Das, and Jason Wei. 2023.
\newblock \href {https://openreview.net/forum?id=fR3wGCk-IXp} {Language models are multilingual chain-of-thought reasoners}.
\newblock In \emph{The Eleventh International Conference on Learning Representations}.

\bibitem[{Shi et~al.(2024)Shi, Xu, Wang, Qin, Wang, Wang, and Wang}]{ShiXWQWWW24}
Haizhou Shi, Zihao Xu, Hengyi Wang, Weiyi Qin, Wenyuan Wang, Yibin Wang, and Hao Wang. 2024.
\newblock \href {https://doi.org/10.48550/ARXIV.2404.16789} {Continual learning of large language models: {A} comprehensive survey}.
\newblock \emph{CoRR}, abs/2404.16789.

\bibitem[{Shoemake(1985)}]{Shoemake85}
Ken Shoemake. 1985.
\newblock \href {https://doi.org/10.1145/325334.325242} {Animating rotation with quaternion curves}.
\newblock In \emph{Proceedings of the 12th Annual Conference on Computer Graphics and Interactive Techniques, {SIGGRAPH} 1985, San Francisco, California, USA, July 22-26, 1985}, pages 245--254. {ACM}.

\bibitem[{Stoica et~al.(2023)Stoica, Bolya, Bjorner, Hearn, and Hoffman}]{StoicaBBHH23}
George Stoica, Daniel Bolya, Jakob Bjorner, Taylor Hearn, and Judy Hoffman. 2023.
\newblock \href {https://doi.org/10.48550/ARXIV.2305.03053} {Zipit! merging models from different tasks without training}.
\newblock \emph{CoRR}, abs/2305.03053.

\bibitem[{Teknium(2023)}]{OpenHermes}
Teknium. 2023.
\newblock \href {https://huggingface.co/datasets/teknium/OpenHermes-2.5} {Openhermes 2.5: An open dataset of synthetic data for generalist llm assistants}.

\bibitem[{Tirumala et~al.(2023)Tirumala, Simig, Aghajanyan, and Morcos}]{TirumalaSAM23}
Kushal Tirumala, Daniel Simig, Armen Aghajanyan, and Ari Morcos. 2023.
\newblock \href {http://papers.nips.cc/paper\_files/paper/2023/hash/a8f8cbd7f7a5fb2c837e578c75e5b615-Abstract-Datasets\_and\_Benchmarks.html} {{D4:} improving {LLM} pretraining via document de-duplication and diversification}.
\newblock In \emph{Advances in Neural Information Processing Systems 36: Annual Conference on Neural Information Processing Systems 2023, NeurIPS 2023, New Orleans, LA, USA, December 10 - 16, 2023}.

\bibitem[{Together.ai(2023)}]{together2023redpajama}
Together.ai. 2023.
\newblock \href {https://github.com/togethercomputer/RedPajama-Data} {Redpajama: an open dataset for training large language models}.

\bibitem[{Touvron et~al.(2023)Touvron, Lavril, Izacard, Martinet, Lachaux, Lacroix, Rozi{\`{e}}re, Goyal, Hambro, Azhar, Rodriguez, Joulin, Grave, and Lample}]{TouvronLIMLLRGHARJGL23}
Hugo Touvron, Thibaut Lavril, Gautier Izacard, Xavier Martinet, Marie{-}Anne Lachaux, Timoth{\'{e}}e Lacroix, Baptiste Rozi{\`{e}}re, Naman Goyal, Eric Hambro, Faisal Azhar, Aur{\'{e}}lien Rodriguez, Armand Joulin, Edouard Grave, and Guillaume Lample. 2023.
\newblock \href {https://doi.org/10.48550/ARXIV.2302.13971} {Llama: Open and efficient foundation language models}.
\newblock \emph{CoRR}, abs/2302.13971.

\bibitem[{V{\'a}radi et~al.(2022)V{\'a}radi, Ny{\'e}ki, Koeva, Tadi{\'c}, {\v{S}}tefanec, Ogrodniczuk, Nito{\'n}, P{\k{e}}zik, Barbu~Mititelu, Irimia, Mitrofan, Tufi{\textcommabelow{s}}, Garab{\'\i}k, Krek, and Repar}]{curlicat}
Tam{\'a}s V{\'a}radi, Bence Ny{\'e}ki, Svetla Koeva, Marko Tadi{\'c}, Vanja {\v{S}}tefanec, Maciej Ogrodniczuk, Bart{\l}omiej Nito{\'n}, Piotr P{\k{e}}zik, Verginica Barbu~Mititelu, Elena Irimia, Maria Mitrofan, Dan Tufi{\textcommabelow{s}}, Radovan Garab{\'\i}k, Simon Krek, and Andra{\v{z}} Repar. 2022.
\newblock \href {https://aclanthology.org/2022.lrec-1.11} {Introducing the {CURLICAT} corpora: Seven-language domain specific annotated corpora from curated sources}.
\newblock In \emph{Proceedings of the Thirteenth Language Resources and Evaluation Conference}, pages 100--108, Marseille, France. European Language Resources Association.

\bibitem[{Winata et~al.(2023)Winata, Xie, Radhakrishnan, Wu, Jin, Cheng, Kulkarni, and Preotiuc{-}Pietro}]{WinataXRWJ0KP23}
Genta~Indra Winata, Lingjue Xie, Karthik Radhakrishnan, Shijie Wu, Xisen Jin, Pengxiang Cheng, Mayank Kulkarni, and Daniel Preotiuc{-}Pietro. 2023.
\newblock \href {https://doi.org/10.18653/V1/2023.FINDINGS-ACL.48} {Overcoming catastrophic forgetting in massively multilingual continual learning}.
\newblock In \emph{Findings of the Association for Computational Linguistics: {ACL} 2023, Toronto, Canada, July 9-14, 2023}, pages 768--777. Association for Computational Linguistics.

\bibitem[{Wolf et~al.(2019)Wolf, Debut, Sanh, Chaumond, Delangue, Moi, Cistac, Rault, Louf, Funtowicz, and Brew}]{WolfDSCD19}
Thomas Wolf, Lysandre Debut, Victor Sanh, Julien Chaumond, Clement Delangue, Anthony Moi, Pierric Cistac, Tim Rault, R{\'{e}}mi Louf, Morgan Funtowicz, and Jamie Brew. 2019.
\newblock \href {https://arxiv.org/abs/1910.03771} {Huggingface's transformers: State-of-the-art natural language processing}.
\newblock \emph{CoRR}, abs/1910.03771.

\bibitem[{Wortsman et~al.(2023)Wortsman, Gururangan, Li, Farhadi, Schmidt, Rabbat, and Morcos}]{lofi}
Mitchell Wortsman, Suchin Gururangan, Shen Li, Ali Farhadi, Ludwig Schmidt, Michael Rabbat, and Ari~S. Morcos. 2023.
\newblock \href {https://openreview.net/forum?id=1U0aPkBVz0} {lo-fi: distributed fine-tuning without communication}.
\newblock \emph{Transactions on Machine Learning Research}.

\bibitem[{Wortsman et~al.(2022)Wortsman, Ilharco, Gadre, Roelofs, Lopes, Morcos, Namkoong, Farhadi, Carmon, Kornblith, and Schmidt}]{WortsmanIGRLMNF22}
Mitchell Wortsman, Gabriel Ilharco, Samir~Yitzhak Gadre, Rebecca Roelofs, Raphael~Gontijo Lopes, Ari~S. Morcos, Hongseok Namkoong, Ali Farhadi, Yair Carmon, Simon Kornblith, and Ludwig Schmidt. 2022.
\newblock \href {https://proceedings.mlr.press/v162/wortsman22a.html} {Model soups: averaging weights of multiple fine-tuned models improves accuracy without increasing inference time}.
\newblock In \emph{International Conference on Machine Learning, {ICML} 2022, 17-23 July 2022, Baltimore, Maryland, {USA}}, volume 162 of \emph{Proceedings of Machine Learning Research}, pages 23965--23998. {PMLR}.

\bibitem[{Xie et~al.(2023)Xie, Pham, Dong, Du, Liu, Lu, Liang, Le, Ma, and Yu}]{Xie0DDLLLL0Y23}
Sang~Michael Xie, Hieu Pham, Xuanyi Dong, Nan Du, Hanxiao Liu, Yifeng Lu, Percy Liang, Quoc~V. Le, Tengyu Ma, and Adams~Wei Yu. 2023.
\newblock \href {http://papers.nips.cc/paper\_files/paper/2023/hash/dcba6be91359358c2355cd920da3fcbd-Abstract-Conference.html} {Doremi: Optimizing data mixtures speeds up language model pretraining}.
\newblock In \emph{Advances in Neural Information Processing Systems 36: Annual Conference on Neural Information Processing Systems 2023, NeurIPS 2023, New Orleans, LA, USA, December 10 - 16, 2023}.

\bibitem[{Xue et~al.(2021)Xue, Constant, Roberts, Kale, Al-Rfou, Siddhant, Barua, and Raffel}]{mt5}
Linting Xue, Noah Constant, Adam Roberts, Mihir Kale, Rami Al-Rfou, Aditya Siddhant, Aditya Barua, and Colin Raffel. 2021.
\newblock \href {https://doi.org/10.18653/v1/2021.naacl-main.41} {m{T}5: A massively multilingual pre-trained text-to-text transformer}.
\newblock In \emph{Proceedings of the 2021 Conference of the North American Chapter of the Association for Computational Linguistics: Human Language Technologies}, pages 483--498, Online. Association for Computational Linguistics.

\bibitem[{Yadav et~al.(2023)Yadav, Tam, Choshen, Raffel, and Bansal}]{YadavTCRB23}
Prateek Yadav, Derek Tam, Leshem Choshen, Colin~A. Raffel, and Mohit Bansal. 2023.
\newblock \href {http://papers.nips.cc/paper\_files/paper/2023/hash/1644c9af28ab7916874f6fd6228a9bcf-Abstract-Conference.html} {Ties-merging: Resolving interference when merging models}.
\newblock In \emph{Advances in Neural Information Processing Systems 36: Annual Conference on Neural Information Processing Systems 2023, NeurIPS 2023, New Orleans, LA, USA, December 10 - 16, 2023}.

\bibitem[{Yang et~al.(2019)Yang, Zhang, Tar, and Baldridge}]{pawsx}
Yinfei Yang, Yuan Zhang, Chris Tar, and Jason Baldridge. 2019.
\newblock \href {https://doi.org/10.18653/v1/D19-1382} {{PAWS}-{X}: A cross-lingual adversarial dataset for paraphrase identification}.
\newblock In \emph{Proceedings of the 2019 Conference on Empirical Methods in Natural Language Processing and the 9th International Joint Conference on Natural Language Processing (EMNLP-IJCNLP)}, pages 3687--3692, Hong Kong, China. Association for Computational Linguistics.

\bibitem[{Yu et~al.(2023)Yu, Yu, Yu, Huang, and Li}]{YuYYHL23}
Le~Yu, Bowen Yu, Haiyang Yu, Fei Huang, and Yongbin Li. 2023.
\newblock \href {https://doi.org/10.48550/ARXIV.2311.03099} {Language models are super mario: Absorbing abilities from homologous models as a free lunch}.
\newblock \emph{CoRR}, abs/2311.03099.

\bibitem[{Yu et~al.(2024)Yu, Jiang, Shi, YU, Liu, Zhang, Kwok, Li, Weller, and Liu}]{metamath}
Longhui Yu, Weisen Jiang, Han Shi, Jincheng YU, Zhengying Liu, Yu~Zhang, James Kwok, Zhenguo Li, Adrian Weller, and Weiyang Liu. 2024.
\newblock \href {https://openreview.net/forum?id=N8N0hgNDRt} {Metamath: Bootstrap your own mathematical questions for large language models}.
\newblock In \emph{The Twelfth International Conference on Learning Representations}.

\bibitem[{Zellers et~al.(2019)Zellers, Holtzman, Bisk, Farhadi, and Choi}]{hellaswag}
Rowan Zellers, Ari Holtzman, Yonatan Bisk, Ali Farhadi, and Yejin Choi. 2019.
\newblock \href {https://doi.org/10.18653/v1/P19-1472} {{H}ella{S}wag: Can a machine really finish your sentence?}
\newblock In \emph{Proceedings of the 57th Annual Meeting of the Association for Computational Linguistics}, pages 4791--4800, Florence, Italy. Association for Computational Linguistics.

\bibitem[{Zhai et~al.(2023)Zhai, Tong, Li, Cai, Qu, Lee, and Ma}]{ZhaiTLCQLM23}
Yuexiang Zhai, Shengbang Tong, Xiao Li, Mu~Cai, Qing Qu, Yong~Jae Lee, and Yi~Ma. 2023.
\newblock \href {https://doi.org/10.48550/ARXIV.2309.10313} {Investigating the catastrophic forgetting in multimodal large language models}.
\newblock \emph{CoRR}, abs/2309.10313.

\bibitem[{Zhang et~al.(2023)Zhang, Fang, Chen, and Namazi{-}Rad}]{ZhangF0N23}
Zihan Zhang, Meng Fang, Ling Chen, and Mohammad{-}Reza Namazi{-}Rad. 2023.
\newblock \href {https://doi.org/10.18653/V1/2023.FINDINGS-EMNLP.633} {{CITB:} {A} benchmark for continual instruction tuning}.
\newblock In \emph{Findings of the Association for Computational Linguistics: {EMNLP} 2023, Singapore, December 6-10, 2023}, pages 9443--9455. Association for Computational Linguistics.

\bibitem[{Zhao et~al.(2024)Zhao, Zhang, Gao, Zhang, Gui, and Huang}]{ZhaoZGZGH24}
Jun Zhao, Zhihao Zhang, Luhui Gao, Qi~Zhang, Tao Gui, and Xuanjing Huang. 2024.
\newblock \href {https://doi.org/10.48550/ARXIV.2401.01055} {Llama beyond english: An empirical study on language capability transfer}.
\newblock \emph{CoRR}, abs/2401.01055.

\end{thebibliography}

\clearpage

\appendix

\section{Extended Evaluation} \label{sec:extended_eval}

\begin{table*}[tbhp]
\centering
\caption{English Benchmark performance of \llama[3][8B] continuously pretrained on Bulgarian}
  \label{tab:cpt_eng}
\resizebox{\linewidth}{!}{
\begin{tabular}{lx{2}{2}x{2}{2}x{2}{2}x{2}{2}x{2}{2}x{2}{2}x{2}{2}x{2}{2}x{2}{2} @{\hskip 10mm}c}
    \toprule
    Training& {WG} & {HS} & {ARC-c} & {ARC-e} & {MMLU} & {Bele} & {MathQA} & {GSM8K} & {TrQA} & {AVG} \\
    \midrule
    Base & 73.00 & 79.12 & $\textbf{53.32}$ & $\textbf{77.65}$ & $\textbf{65.16}$ & 66.77 & 40.00 & 47.99 & $\textbf{71.62}$ & 63.85 \\
    \cmidrule(lr){1-1}
    CPT & 74.43 & 80.00 & 50.34 & 71.59 & 61.84 & $\textbf{71.33}$ & 38.29 &  $\textbf{71.95}$ & 63.79 & 64.84 \\
    \method & $\mbf{74.58}$ & $\mbf{80.12}$ & 52.81 & 75.54 & 63.25 & 71.11 & $\mbf{41.07}$ & 69.97 & 67.65 & $\mbf{66.23}$ \\
    \bottomrule
    \end{tabular}
}
\end{table*}

\begin{table*}[tbhp]
\centering
\caption{English Benchmark performance of \llama[3][8B] continuously pretrained on German}
  \label{tab:cpt_de_eng}
\resizebox{\linewidth}{!}{
\begin{tabular}{lccccccccc @{\hskip 10mm}c}
    \toprule
    Model& WG & HS & ARC-c & ARC-e & MMLU & Bele & MathQA & GSM8K & TrQA&AVG \\
    \midrule
    Base & $\textbf{73.00}$ & $\textbf{79.12}$ & $\textbf{53.32}$ & $\textbf{77.65}$ & $\textbf{65.16}$ & $\textbf{66.77}$ & $\textbf{40.00}$ & $\textbf{47.99}$ & $\textbf{71.62}$ & $\textbf{63.85}$ \\
    \cmidrule(lr){1-1}
    CPT& $72.45$ & ${78.80}$ & ${51.19}$ & ${77.60}$ & $62.90$ & $55.88$ & $39.27$ &  $41.16$ & $67.78$ & $60.79$ \\

    \method& $72.84$ & $78.65$ & $50.85$ & $77.02$ & ${63.87}$ & ${58.66}$ & ${39.83}$ & ${44.73}$ & ${69.23}$ & ${61.74}$ \\
    \bottomrule
    \end{tabular}
}

\end{table*}

\begin{table}[t]
\centering
\caption{Effect of data slice order on \method.}
  \label{tab:slice_order}
\resizebox{0.9\linewidth}{!}{
\begin{tabular}{l
                x{2}{2}
                x{2}{2}
                x{2}{3}
                x{1}{3}}
    \toprule
Data Order & {Avg EN} & {Avg BG} & {BG pplx.} & {EN pplx.}\\
\midrule
    Base Model        & 63.8525183 & 44.18281907 &1.695 &2.042 \\
    \cmidrule(lr){1-1}
    Standard & \textbf{66.23} & 53.06475341 & \textbf{1.061} &2.09731 \\
    Reversed & 65.64 & 52.70 & 1.16723 & 2.07083 \\
    \cmidrule(lr){1-1}
    Merged & 66.26  & \textbf{53.34} & 1.07601 &  \textbf{2.069}\\
\bottomrule
\end{tabular}
}
\end{table}
\paragraph{Data Slice Order}
We evaluate the effect of data slice order on \llama[3] in \cref{tab:slice_order}. We observe that reversing the order of data slices for \method training has only a minimal effect on both forgetting and domain adaptation. Interestingly, merging the two models obtained via these different orders is strictly better than either, although at twice the computational cost. This highlights again the effectiveness of \method at finding optimal task vectors by merging out the error component.

\begin{table}[t]
    \centering
    \caption{Effect of tokenizer extension on performance before and after continuous pertaining (CPT) of \mistral[7B].}
    \vspace{-2mm}
    \label{tab:tok_extension}
    \resizebox{0.95\linewidth}{!}{
    \begin{tabular}{llccc}
         \toprule
         Training & Extension & Fertility & Avg BG & Avg EN \\
         \midrule
         \multirow{2}{*}{Base} &None & 2.37 & 44.50 & 63.50 \\
         &$8$k & \textbf{1.71} & 29.28 & 62.57 \\
         \cmidrule(lr){1-2}
         \multirow{2}{*}{CPT} & None & 2.37 & \textbf{51.47} & \textbf{61.48} \\
         & $8$k & \textbf{1.71} & 50.93 & 60.96 \\
         \bottomrule
    \end{tabular}
    }
\end{table}

\paragraph{Tokenizer Extension}
A common challenge with LLM domain adaptation is that the LLM's tokenizer may not be well suited for the target domain, expressed in a higher fertility. This entails longer training, slower inference, shorter effective context length as well as potential performance degradation.

In our language transfer setting from English to Bulgarian, we also make use of tokenizer/vocabulary expansion for some of our experiments to reduce the computational cost. In the case of \mistral[7B], we find that Bulgarian tokenization is subpar. To this end, we train a SentencePieceBPE \citep{sentencepiece} tokenizer with a vocabulary of $8$k tokens on high-quality Bulgarian text. We find that a mix of 75\% RPv2 BG and 25\% Wikipedia, where the whole Bulgarian Wikipedia comprises these 25\% gave the lowest fertility on a sample from mC4 \citep{mt5}. After removing all tokens that do not include at least one Cyrillic character or are already in the original tokenizer, we are left with exactly 6000 new tokens, which are then appended to the original Mistral tokenizer with their respective SentencePiece scores. This whole procedure ensures that the English tokenization remains practically unchanged, which is important to reduce Catastrophic Forgetting.
We initialize the new input and output embeddings with their mean tokenization using the original tokenizer and add them to the model's vocabulary in the style of VIPI \citep{vipi} and FVT \citep{fvt}. We report results for \mistral[7B] in \cref{tab:tok_extension} and use an $8$k (effectively $6$k) tokenizer extension for all further \mistral[7B] experiments due to the greatly increased training and inference efficiency at very similar performance and retain the original \llama[3][8B] tokenizer due to its already huge vocabulary and lower fertility.
Note: Reducing or increasing the amount of Web data in that tokenizer training mix resulted in higher fertility on the mC4 sample. The reason for this is not fully clear and we intend on investigating this in future work.

\paragraph{Low-Rank Adaptation}

LoRA has become widely popular as a method for cheaper finetuning of LLMs \citep{hu2022lora}. Taking into consideration the contribution of \citep{biderman2024lora}, which puts LoRA in the context of learning less but also forgetting less we also show how LoRA fairs in our Language Transfer setting. Due to limited compute resources, we do not perform an extensive hyperparameter sweep and instead copy what we can from the Code CPT experiment in \citet{biderman2024lora}. As far as we know the batch sizes are not mentioned there and we decide to stick to $512$, while deducting that the original may have used $128$. We also proportionately increase the learning rate and find that $4e-5$ converges the fastest. The comparison in Table \ref{tab:lora} is in the reduced, $20$B-token setting, same as in Table \ref{tab:parallelism}. We indeed observe a better preservation of the English Negative Log Likelihood but also a significant reduction in learned Bulgarian capabilities. It may be the case that the Language Transfer adaptation is not as low-rank as it is for Code and the referred LoRA rank parameter should be set higher than $256$. 

\begin{table}[t]
    \centering
    \caption{Effect of \lora regularization compared to \method on \llama[3].}
    \vspace{-2mm}
    \label{tab:lora}
    \resizebox{0.8\linewidth}{!}{
    \begin{tabular}{lcccc}
         \toprule
         Training & Avg BG & Avg EN & BG nll & EN nll \\
         \midrule
         Base & 44.18 & 63.85 & 1.695 & 2.042 \\
         \cmidrule(lr){1-1}
         CPT & 51.76 & 66.33 & \textbf{1.136} & 2.093\\
         \method & \textbf{52.01} & \textbf{67.00} & 1.194 & \textbf{2.077}\\
         \lora & 45.33 & 64.71 & 1.515 & 2.059\\
         \bottomrule
    \end{tabular}
    }
\end{table}

\section{Benchmark Details}

\subsection{Benchmark Descriptions} \label{sec:app_benchmarks}
Below we provide short descriptions of all datasets and note the license they are published under. German language benchmarks is run in a 5-shot setting. For the other evaluations, we specify the number of shots below, or use 0-shots when not specified.

\paragraph{HellaSwag} (MIT License) \citep{hellaswag} is a common sense reasoning benchmark asking an LLM to select a logical continuation of a sentence. Evaluated on the 10000 sample validation set.

\paragraph{Winogrande} (Appache 2.0 License)\citep{winogrande} is a common sense reasoning benchmark asking an LLM to fill in a blank from a choice of two entities to logically complete a sentence. Evaluated on the 1767 sample validation set of \texttt{winogrande\_xl}.

\paragraph{ARC-Easy and -Challenge} (CC BY-SA License) \citep{arc} is a dataset of science exam questions. Evaluated on the 2590 hard sample (ARC-Challenge) and 5197 easy samples (ARC-Easy).

\paragraph{MMLU} (MIT License) \citep{mmlu} is a multitask language understanding benchmark covering a wide range of 57 different tasks. Evaluated on 14079 test set samples. We evaluate MMLU using 5-shots.

\paragraph{GSM8K} (MIT License) \citep{gsm8k} is a mathematical reasoning benchmark consisting of grade-school math questions for which free text answers must be provided. Evaluated on 1.3k test set samples. We run GSM8k with 8-shot chain-of-thought generation.

\paragraph{MathQA} (Apache 2.0 License) \citep{mathqa} is a multiple choice mathematical reasoning benchmark. Evaluated on 4475 validation set samples.

\paragraph{Belebele} (CC-BY-NC 4.0 License) \citep{belebele} is a multiple choice reading comprehension dataset. Evaluated on 900 samples per language.

\paragraph{TriviaQA} (Apache 2.0 License) \citep{triviaqa}) is trivia question dataset. Evaluated on 17.9k validation set samples. We use 5-shot evaluation.

\paragraph{XNLI} (CC BY-NC 4.0 License) \citep{xnli} is a language understanding dataset where the task is to decide whether two statements contradict one-another, are neutral, or one entails the other. Evaluated on 2.5k validation samples.

\paragraph{EXAMS} (CC BY-SA 4.0 License) \citep{exams} is a high school exam question dataset covering a range of subjects. Evaluated 1472 test set samples in Bulgarian. We use 5-shot evaluation.

\paragraph{PAWS} (Special License permitting "free use for any purpose") \citep{pawsx} is a reading comprehension dataset where the task is to decide whether two benchmarks are paraphrases. Evaluated on 1967 test set samples.

\paragraph{MGSM} (CC BY 4.0 License) \citep{mgsm} is mathematical reasoning benchmark manually translated from GSM8k. Evaluated on 250 test set samples.

\subsection{Bulgarian Benchmarks} \label{sec:translation}
\paragraph{Translation}
We use Google Translate to machine translate the text of the benchmark problems and answers. Additional, we identified a set of heuristics for cases where the machine translation is of low quality, such as inconsistent translations of the same word and not following exact format in both source and target sentences. In all such cases, we gave the tasks to human translators with additional instructions on possible problems we identified in each benchmark. Overall human translators manually translated 2143 test set samples.

A notable example of a benchmark with significant problems that we expect to repeat in many other languages is Winogrande challenge~\citep{winogrande}. In this case, one of two words have to be chosen based on world knowledge and reasoning. However, with machine translation or na\"ive human translation to non-English, the actual answer can be revealed in a much easier way by the means of having only one answer that is in gender agreement with other words in the sentence. We performed manual translations that used synonyms that do not exhibit such behavior and as a result, the translated benchmark is not easier than the original. The translated versions of the benchmarks with these fixes are made publicly available.

\paragraph{MON} The MON dataset is obtained as private data from the Bulgarian Ministry of Education. This contains 10088 exam questions with 4 possible choices, only one of which is correct, spanning topics from 4th to 12th grade tests previously given for external tests to schools in Bulgaria. The questions span all subjects tested by the official Bulgarian curriculum but exclude problems such as geometry tasks that include images in their problem definition or answers. The dataset is not publicly available and as a result, we expect it to be less likely to be in any of the training data in any form.

\section{Dataset Details}\label{sec:app_data}

\subsection{IFT Set Composition}
We make note of the good performance and instruction following capabilities of the Intel Neural-Chat models and decide to include SlimOrca \citep{slimorca,mukherjee2023orca,theflancollection} and MetaMathQA \citep{metamath} in our English IFT data mix. To fill in the gap of multi-turn conversation data we additionally include the Capybara dataset \citep{capybara}, which we have observed from our experience boost the models' "chattiness" and overall response quality.

The fact that there are no publicly available general Bulgarian IFT datasets, lead us to the translation of already existing ones. We use machine translation to produce $50$K Bulgarian translated samples from the OpenHermes-2.5 \citep{OpenHermes} dataset, $10$K samples from MetaMathQA \citep{metamath} and $2$K samples of code with Bulgarian instructions from CodeAlpaca \citep{codealpaca}. We take special care in the translation of the Capybara \citep{capybara} and OpenHermes datasets. Through a combination of classification and manual inspection, we identify examples, where the machine translation is not good enough to make a sensible training example, e.g. instructions that require rhyming, as the words that rhyme in English will most likely not rhyme in Bulgarian. The identified 5\% of the Capybara dataset is then manually translated/adjusted to fit the Bulgarian language.  See \cref{tab:data_bg_lic} for full details and licenses.
\begin{table}[t]
\centering
\caption{Composition of the Bulgarian IFT dataset.}
\label{tab:bg_ift_data}
\resizebox{\linewidth}{!}{
\begin{tabular}{llccc}
    \toprule
    Dataset & Domain & \#Examples & Repetitions &  Prob [\%] \\
    \midrule
    OpenHermes-2.5-BG & Mixed Conversations&  $50,000$ & $1$ & $64.10$ \\
    Capybara-BG & Mixed Conversations& $16,000$ & $1$ & $20.51$\\
    MetaMath-BG & Math & $10,000$ & $1$ & $12.82$\\
    CodeAlpaca-BG & Code & $2,000$ & $1$ & $2.56$\\
    \bottomrule
    \end{tabular}
}
\end{table}

\begin{table}[t]
\centering
\caption{Composition of the English IFT dataset.}
\label{tab:en_ift_data}
\resizebox{\linewidth}{!}{
\begin{tabular}{llccc}
    \toprule
    Dataset & Domain & \#Examples & Repetitions & Prob [\%] \\
    \midrule
    SlimOrca & Mixed Conversations&  $517,982$ & $1$ & $55.76$ \\
    MetaMathQA & Math & $395,000$ & $1$ & $42.52$\\
    Capybara & Mixed Conversations& $16,000$ & $1$ & $1.72$\\
    \bottomrule
    \end{tabular}
}
\end{table}

\begin{table}[t]
\centering
\caption{Composition of the German IFT dataset.}
\label{tab:de_ift_data}
\resizebox{\linewidth}{!}{
\begin{tabular}{llccc}
    \toprule
    Dataset & Domain & \#Examples & Repetitions &  Prob [\%] \\
    \midrule
    evol-instruct-deutsch  & Mixed Conversations & $59,022$ & $1$ & $45.15$\\
    alpaca-gpt4-deutsch& Mixed Conversations&  $50,000$ & $1$ & $38.23$ \\
    OpenSchnabeltier & Mixed Single-turn& $21,749$ & $1$ & $16.62$\\
    \bottomrule
    \end{tabular}
}
\end{table}

\begin{table}[t]
\centering
\caption{Sources and licenses of used datasets}
\vspace{-2mm}
\resizebox{\linewidth}{!}{
\begin{tabular}{llc}
    \toprule
    Dataset & Source & License  \\
    \midrule
    RPv2 pipeline & \citet{together2023redpajama}  & Apache 2.0 \\
    OpenWebText & \citet{openwebtext} & CC0-1.0   \\
    CulturaX & \citet{culturax} & CC0-1.0   \\
    FineWeb-Edu & \citet{fineweb-edu}& ODC-BY \\
    PubMed & \citet{pubmed}& Unknown\\
    Eur-Lex & \citet{baisa-etal-2016-european} & CC-BY-NC-SA   \\
    Wikipedia & \citet{wikidump} & CC-BY-SA-3.0  \\
    OrcaMath & \citet{mitra2024orcamath} & MIT  \\
    Parlamint & \citet{parlamint} & CC-BY  \\
    OpenHermes-2.5 & \citet{OpenHermes}& Unknown\\
    Capybara & \citet{capybara} & Apache 2.0 \\
    Curlicat & \citet{curlicat} & CC-BY-SA-4.0  \\
    SlimOrca & \citet{slimorca,mukherjee2023orca} & MIT  \\
    CodeAlpaca & \citet{codealpaca} & CC-BY-4.0  \\
    Europarl & \citet{koehn-2005-europarl} & Unknown \\
    MetaMath & \citet{metamath} & MIT  \\
    Open-Platypus & \citet{platypus2023} & Apache 2.0  \\
    alpaca-gpt4-deutsch & \citet{Chen_MultilingualSIFT_Multilingual_Supervised_2023} & Apache 2.0 \\
    OpenSchnabeltier & \citet{openschnabeltier} & Apache 2.0\\
    evol-instruct-deutsch & \citet{Chen_MultilingualSIFT_Multilingual_Supervised_2023} & Apache 2.0\\
    \bottomrule
    \end{tabular}
}
\label{tab:data_bg_lic}
\vspace{-3mm}
\end{table}

\begin{table}[t]
\centering
\caption{Composition of the Bulgarian curriculum splits.}
\label{tab:curriculum}
\resizebox{\linewidth}{!}{
\begin{tabular}{lcclcc}
    \toprule
    Split & \# Total BG & \# Total Replay & Dataset & Repetitions & Replay \\
    \midrule
    \multirow{6}{*}{$\bc{X}_1$} & \multirow{6}{*}{14.7B}& \multirow{6}{*}{850M}& Wikipedia-BG & 1 & \xmark\\
    &&&OpenWebText & 0.1 & \cmark\\
    &&&Bulgarian Law & 1 & \xmark\\
    &&&Eur-Lex-BG & 1 & \xmark\\
    &&&IFT-BG & 1 & \xmark\\
    &&&RPv2-BG & 0.2 & \xmark\\
    \cmidrule(lr){1-3}
    \multirow{5}{*}{$\bc{X}_2$} & \multirow{5}{*}{8.3B}& \multirow{5}{*}{3.3B}& Wikipedia-EN & 0.25 & \cmark\\
    &&&OpenWebText & 0.15 & \cmark\\
    &&&GitHub repos & 0.2 & \cmark\\
    &&&IFT-EN & 1 & \cmark\\
    &&&RPv2-BG & 0.12 & \xmark\\
    \cmidrule(lr){1-3}
    \multirow{9}{*}{$\bc{X}_3$} & \multirow{9}{*}{11.4B}& \multirow{9}{*}{850M}& Wikipedia-BG & 1 & \xmark\\
    &&&OpenWebText & 0.1 & \cmark\\
    &&&Bulgarian Law & 1 & \xmark\\
    &&&Eur-Lex-BG & 1 & \xmark\\
    &&&IFT-BG & 1 & \xmark\\
    &&&RPv2-BG & 0.12 & \xmark\\
    &&&Parlamint-BG & 1 & \xmark\\
    &&&Europarl-BG & 1 & \xmark\\
    &&&Legal docs & 0.4 & \xmark\\
    \cmidrule(lr){1-3}
    \multirow{5}{*}{$\bc{X}_4$} & \multirow{5}{*}{8.3B}& \multirow{5}{*}{3.3B}& Wikipedia-EN & 0.25 & \cmark\\
    &&&OpenWebText & 0.15 & \cmark\\
    &&&GitHub repos & 0.2 & \cmark\\
    &&&IFT-EN & 1 & \cmark\\
    &&&RPv2-BG & 0.12 & \xmark\\
    \cmidrule(lr){1-3}
    \multirow{9}{*}{$\bc{X}_5$} & \multirow{9}{*}{12.4B}& \multirow{9}{*}{850M}& Wikipedia-BG & 1 & \xmark\\
    &&&OpenWebText & 0.1 & \cmark\\
    &&&Bulgarian Law & 1 & \xmark\\
    &&&Books & 1 & \xmark\\
    &&&IFT-BG & 1 & \xmark\\
    &&&RPv2-BG & 0.1 & \xmark\\
    &&&Parlamint-BG & 1 & \xmark\\
    &&&Europarl-BG & 1 & \xmark\\
    &&&Legal docs & 0.4 & \xmark\\
    \cmidrule(lr){1-3}
    \multirow{5}{*}{$\bc{X}_6$} & \multirow{5}{*}{8.3B}& \multirow{5}{*}{3.3B}& Wikipedia-EN & 0.25 & \cmark\\
    &&&OpenWebText & 0.15 & \cmark\\
    &&&GitHub repos & 0.2 & \cmark\\
    &&&IFT-EN & 1 & \cmark\\
    &&&RPv2-BG & 0.12 & \xmark\\
    \cmidrule(lr){1-3}
    \multirow{9}{*}{$\bc{X}_7$} & \multirow{9}{*}{10.3B}& \multirow{9}{*}{850M}& Wikipedia-BG & 1 & \xmark\\
    &&&OpenWebText & 0.1 & \cmark\\
    &&&Bulgarian Law & 1 & \xmark\\
    &&&Books & 1 & \xmark\\
    &&&IFT-BG & 1 & \xmark\\
    &&&RPv2-BG & 0.1 & \xmark\\
    &&&Parlamint-BG & 1 & \xmark\\
    &&&Europarl-BG & 1 & \xmark\\
    &&&Legal docs & 0.2 & \xmark\\
    \cmidrule(lr){1-3}
    \multirow{5}{*}{$\bc{X}_8$} & \multirow{5}{*}{8.3B}& \multirow{5}{*}{3.7B}& Wikipedia-EN & 0.25 & \cmark\\
    &&&OpenWebText & 0.15 & \cmark\\
    &&&GitHub repos & 0.4 & \cmark\\
    &&&IFT-EN & 1 & \cmark\\
    &&&RPv2-BG & 0.12 & \xmark\\
    \bottomrule
    \end{tabular}
}
\end{table}

\subsection{Validation Set Composition}
Constructing validation datasets for language model training, especially when such are trained on web-crawl data, is a challenging task with respect to avoiding data contamination. Our Bulgarian validation set consists of a total of $40$K examples, $30$K of which are a held-out set of news articles from a specific media outlet and the other $10$K is a mix of dialogs, questions and answers, literary works and legal documents. The English validation dataset is comprised of $25$K random samples from the FineWeb-Edu dataset \citep{fineweb-edu} $7$K samples from arXiv scientific papers, $3$K from the PubMed dataset \citep{pubmed} and $5$K books from the Project Gutenberg\footnote{\url{https://www.gutenberg.org/}}.

\section{Experimental Setup and Evaliation Details}
\subsection{Training parameters}
We use the same exact training hyperparameters for both \mistral[7B] and \llama[3][8B] based models.
We stick to the 8192 size context lengths and train with sequence packing, without truncation.
Based on prior work and initial experiments, we find that $1e-5$ is the best maximum learning rate for continued pre-training in our settings with a batch size of $512$ for continued pre-training and $256$ for supervised fine-tuning, effectively training for $4$M and $2$M tokens respectively.
The optimizer in use is AdamW with $\beta_1=0.9$ and $\beta_2=0.95$ and $0.05$ weight decay rate. We use a cosine decay learning rate scheduler, that decays the LR to $0.1 \cdot \text{max\_lr}$ with $\max(100,0.01\cdot\text{total\_steps})$ of linear warmup.

For fine-tuning, we have found that training for more than 2 epochs on a given IFT dataset with the aforementioned hyperparameters is not beneficial and exaggerates catastrophic forgetting. Additionally, we add embedding vector noise during training through NEFTune \citep{jain2024neftune} with a noise-$\alpha=5$. In this stage, we train only on the IFT completions and not on the prompts. This is important to prevent unwanted self-talking behavior in live usage.

Since we train on 64 GPUs at once, we exploit DeepSpeed ZeRO \citep{deepspeed10.1145/3394486.3406703,dszero} stage 1 with mixed precision training in bf16. Combining this with activation checkpointing and FlashAttention-2 \citep{dao2024flashattention} allows us to use a batch size of 2 during training and evaluation. For reference, our setup allows the models to train with up to 7000 tokens per second per GPU.
\subsection{Computational Budget}
All model training and evaluations were conducted on a cluster of 64 NVIDIA H100 GPUS (8 nodes x 8 GPUs) with InfiniBand and 224 available CPU cores per node.
The total computational cost of the experiments included in this paper, including exploratory ones not mentioned here, is around $80,000$ NVIDIA H100 GPU hours. The tokenizer extension we perform on \mistral[7B (Base)] helps reduce the training and inference cost of our Mistral-based models by roughly 30\%.

\end{document}